\documentclass{article}

\usepackage[utf8]{inputenc} % allow utf-8 input
\usepackage[T1]{fontenc}    % use 8-bit T1 fonts
\usepackage{hyperref}       % hyperlinks
\usepackage{url}            % simple URL typesetting
\usepackage{booktabs}       % professional-quality tables
\usepackage{amsfonts}       % blackboard math symbols
\usepackage{nicefrac}       % compact symbols for 1/2, etc.
\usepackage{microtype}      % microtypography
\usepackage{booktabs,caption}
\usepackage[flushleft]{threeparttable}
\usepackage{authblk}
\usepackage{fullpage}

\usepackage{graphicx,color}
\usepackage{epsfig}
\usepackage{amsmath,amsthm,amssymb}
\usepackage{subfigure}
\usepackage{multirow}
\usepackage[linesnumbered,ruled]{algorithm2e}
\newtheorem{thm}{Theorem}
\newtheorem{assumption}{Assumption}

\newtheorem{lem}{Lemma}
\newtheorem{proposition}{Proposition}
\newtheorem{defn}{Definition}

\theoremstyle{remark}
\newtheorem{rem}{Remark}

\newcommand{\N}{\mathbb{N}}

\newcommand{\x}{\mathbf{x}}

\newcommand{\F}{\mathbf{F}}
\newcommand{\w}{\mathbf{w}}
\newcommand{\E}{\mathbb{E}}
\graphicspath{{figures/}}

\title{Collaborative Deep Learning in \\ Fixed Topology Networks}
\author[1]{Zhanhong Jiang}
\author[1]{Aditya Balu}
\author[2]{Chinmay Hegde}
\author[1]{Soumik Sarkar}
\affil[1]{Department of Mechanical Engineering, Iowa State University\\
\(zhjiang, baditya, soumiks\)@iastate.edu}
\affil[2]{Department of Electrical and Computer Engineering , Iowa State University\\
chinmay@iastate.edu}

\begin{document}
%\nipsfinalcopy

\maketitle

\begin{abstract}
There is significant recent interest to parallelize deep learning algorithms in order to handle the enormous growth in data and model sizes. While most advances focus on model parallelization and engaging multiple computing agents via using a central parameter server, aspect of data parallelization along with decentralized computation has not been explored sufficiently. In this context, this paper presents a new consensus-based distributed SGD (CDSGD) (and its momentum variant, CDMSGD) algorithm for collaborative deep learning over fixed topology networks that enables data parallelization as well as decentralized computation. Such a framework can be extremely useful for learning agents with access to only local/private data in a communication constrained environment. We analyze the convergence properties of the proposed algorithm with strongly convex and nonconvex objective functions with fixed and diminishing step sizes using concepts of Lyapunov function construction. We demonstrate the efficacy of our algorithms in comparison with the baseline centralized SGD and the recently proposed federated averaging algorithm (that also enables data parallelism) based on benchmark datasets such as MNIST, CIFAR-10 and CIFAR-100.
%For validating the proposed approaches, we use benchmark datasets, such as CIFAR-10 (results for MNIST and CIFAR-100 are included in the supplementary section~\ref{supp}).
%Stochastic Gradient Descent (SGD) based learning methods are have gained considerable attractions in deep learning areas due to its simple applicability and high reliability for most problems. However, problem scales have been increasing due to more and more data collected such that traditionally centralized SGD may not satisfy the problem requirements. Moreover, for some problems, machines may not get access to the entire training data set because of the hardware condition limitations. Therefore, collaborative learning methodologies corresponding to data parallelism and model parallelism are developed for addressing such issues. This paper proposes consensus-based distributed SGD (CDSGD) methods and
%analyzes convergence properties for strongly convex and non-convex objective functions by establishing Lyapunov function and stochastic Lyapunov gradient. Two benchmark data sets, i.e., CIFAR-10 and CIFAR-100, are used to validate our proposed algorithms with comparison to a recently developed method Federated Averaging (FedAvg).
\end{abstract}

\vspace{-10pt}
\section{Introduction}
%Deep learning has been playing a significantly critical role in artificial intelligence and still keeps growing up with a fast pace under the situation where data analysis is of paramount importance and interest in numerous areas, such as health care, finance, transportation, power management, weather forecast, social networks, etc.~\cite{bottou2016optimization,lecun2015deep}
%%Machine learning provides various effective and powerful methodologies that can find relations within the data while applying them to prediction, forecast, inference, and decision making.
%The current state-of-the-art includes many approaches which rely only on centralized manner machines to train and test data sets associated with problems. When problems become more and more complex with a large amount of data to require more expansive computational resources, centralized manner machines cannot satisfy the requirements in order to obtain promising and convincing results. For example, deep learning applies to solve large-scale problems, such as image or speech recognition~\cite{gupta2015model}, and meanwhile correspondingly comes up with the considerable computational requirements.
In this paper, we address the scalability of optimization algorithms for deep learning in a distributed setting. Scaling up deep learning~\cite{lecun2015deep} is becoming increasingly crucial for large-scale applications where the sizes of both the available data as well as the models are massive~\cite{gupta2015model}. Among various algorithmic advances, many recent attempts have been made to parallelize stochastic gradient descent (SGD) based learning schemes across multiple computing agents. An early approach called Downpour SGD~\cite{dean2012large}, developed within Google's \textit{disbelief} software framework, primarily focuses on model parallelization (i.e., splitting the model across the agents). %Though the framework allows for data parallelization (i.e., splitting the dataset \blue{across} the agents), \red{that aspect was not explored sufficiently.}
A different approach known as elastic averaging SGD (EASGD)~\cite{zhang2015deep} attempts to improve perform multiple SGDs in parallel; this method uses a central parameter server that helps in assimilating parameter updates from the computing agents. However, none of the above approaches concretely address the issue of \emph{data} parallelization, which is an important issue for several learning scenarios: for example, data parallelization enables privacy-preserving learning in scenarios such as distributed learning with a network of mobile and Internet-of-Things (IoT) devices. A recent scheme called \emph{Federated Averaging SGD}~\cite{mcmahan2016communication} attempts such a data parallelization in the context of deep learning with significant success; however, they still use a central parameter server.

In contrast, deep learning with decentralized computation can be achieved via gossip SGD algorithms~\cite{blot2016gossip,jin2016scale}, where agents communicate probabilistically without the aid of a parameter server. However, decentralized computation in the sense of gossip SGD is not feasible in many real life applications. For instance, consider a large (wide-area) sensor network~\cite{MRWG11,LGLPS17} or multi-agent robotic network that aims to learn a model of the environment in a collaborative manner~\cite{CH10,JCSR16}. For such cases, it may be infeasible for arbitrary pairs of agents to communicate on-demand; typically, agents are only able to communicate with their respective neighbors in a communication network in a fixed (or evolving) topology.

\setlength\tabcolsep{2pt}
\begin{table}
\begin{center}
\begin{threeparttable}
\caption{Comparisons between different optimization approaches}\label{table1}
  \begin{tabular}{c c c c c c c c}
    \hline
    Method & $f$ & $\nabla f$ & Step Size & Con.Rate & D.P. & D.C. & C.C.T. \\ \hline
    SGD & Str-con & Lip. & Con.  & $\mathcal{O}(\gamma^k)$ & No & No & No \\
    Downpour SGD~\cite{dean2012large} & Nonconvex & Lip. & Con.\&Ada.  & N/A & Yes & No & No  \\
    EASGD~\cite{zhang2015deep} & Str-con & Lip. & Con. & $\mathcal{O}(\gamma^k)$  & No & No & No\\  \cline{2-8}
    %Gossip SGD & Strongly convex & Lipschitz, bounded & Constant/Diminshing & $\mathcal{O}(\gamma^k)(0\textless\gamma\textless1)$/$\mathcal{O}(\frac{1}{k}$ \\
    \multirow{2}{*}{Gossip SGD~\cite{jin2016scale}} & Str-con & Lip.\&Bou. & Con. & $\mathcal{O}(\gamma^k)$ &\multirow{2}{*} {No} & \multirow{2}{*}{Yes} & \multirow{2}{*}{No}\\
    &Str-con & Lip.\&Bou. & Dim. &$\mathcal{O}(\frac{1}{k}$)\\ \cline{2-8}
    FedAvg~\cite{mcmahan2016communication}& Nonconvex & Lip. & Con. & N/A & Yes & No & No\\ \cline{2-8}
    \multirow{4}{*}{CDSGD [This paper]} & Str-con & Lip.\&Bou. & Con. & $\mathcal{O}(\gamma^k)$ & \multirow{4}{*} {Yes} & \multirow{4}{*}{Yes} & \multirow{4}{*}{Yes}\\
    & Str-con & Lip.\&Bou. & Dim. & $\mathcal{O}(\frac{1}{k^\epsilon})$ \\
    & Nonconvex & Lip.\&Bou. & Con. & N/A \\
    & Nonconvex & Lip.\&Bou. & Dim.& N/A\\
    \hline
  \end{tabular}

\begin{tablenotes}
    \small
    \item Con.Rate: convergence rate, Str-con: strongly convex. Lip.\&Bou.: Lipschitz continuous and bounded. Con.: constant and Con.\&Ada.: constant\&adagrad. Dim.: diminishing. $\gamma\in(0,1)$ is a positive constant. $\epsilon\in(0.5,1]$ is a positive constant. D.P.: data parallelism. D.C.: decentralized computation. C.C.T.: constrained communication topology.
  \end{tablenotes}\vspace{-20pt}
\end{threeparttable}
\end{center}
\end{table}

\vspace{0pt}
\textbf{Contribution}: This paper introduces a new class of approaches for deep learning that enables both data parallelization and decentralized computation. Specifically, we propose consensus-based distributed SGD (CDSGD) and consensus-based distributed momentum SGD (CDMSGD) algorithms for collaborative deep learning that, for the first time, satisfies all three requirements: data parallelization, decentralized computation, and constrained communication over fixed topology networks. Moreover, while most existing studies solely rely on empirical evidence from simulations, we present rigorous convergence analysis for both (strongly) convex and non-convex objective functions, with both fixed and diminishing step sizes using a Lyapunov function construction approach.  Our analysis reveals several advantages of our method: we match the best existing rates of convergence in the centralized setting, while simultaneously supporting data parallelism as well as constrained communication topologies; to our knowledge, this is the first approach that achieves all three desirable properties; see Table~\ref{table1} for a detailed comparison.

Finally, we validate our algorithms' performance on benchmark datasets, such as MNIST, CIFAR-10, and CIFAR-100. Apart from centralized SGD as a baseline, we also compare performance with that of Federated Averaging SGD as it also enables data parallelization. Empirical evidence (for a given number of agents and other hyperparametric conditions) suggests that while our method is slightly slower, we can achieve higher accuracy compared to the best available algorithm (Federated Averaging (FedAvg)).

%presents the detailed comparison between our results and other existing related works in conditions on objective functions and step size, convergence rate, data parallelism, decentralized computation, and constrained communication topology.
%to show the statistical performance and compare with FedAvg,

%which is evidently shown as the state-of-the-art dominant distributed algorithm. In this context, CDSGD enables all local workers (agents) to communicate with each other synchronously and the communication period is 1 as in each iteration every agent is required to communicate with other agents in its neighborhood such that each agent asymptotically converges to the same stationary point (global optimum or local optimum).

%For addressing problems of how to collaboratively learn based on partial information in deep learning, this paper presents the consensus-based distributed SGD (CDSGD) methods which adopt the consensus protocol without identifying a central variable to guide local variables.
%Different from the Federated Averaging (FedAvg)~\cite{mcmahan2016communication} in which a master parameter server are required, CDSGD demands the collaborative learning among workers which use different parts of data to guarantee the asymptotic consensus.

\textbf{Related work}:
%For addressing issues in traditional deep learning problems, distributed algorithms or methodologies have been an attractive technique that gains more and more concerning in the deep learning community which enable multiple machines to cooperatively solve problems. In~\cite{dean2012large} Dean et al proposed and developed Downpour SGD in order to achieve large-scale distributed training.
%For scaling up deep learning algorithms Coates et al~\cite{catanzaro2013deep} presented a cluster of GPU servers with Infini-band interconnects and message passing interface (MPI) based on Commodity Off-The-Shelf High Performance Computing (COTS HPC) technology to solve the large-scale training problems.
Apart from the algorithms mentioned above, a few other related works exist, including a distributed system called Adam for large deep neural network (DNN) models~\cite{chilimbi2014project} and a distributed methodology by Strom~\cite{strom2015scalable} for DNN training by controlling the rate of weight-update to reduce the amount of communication. Natural Gradient Stochastic Gradient Descent (NG-SGD) based on model averaging~\cite{su2015experiments} and staleness-aware async-SGD~\cite{zhang2015staleness} have also been developed for distributed deep learning. A method called CentralVR~\cite{de2016efficient} was proposed for reducing the variance and conducting parallel execution with linear convergence rate. Moreover, a decentralized algorithm based on gossip protocol called the multi-step dual accelerated (MSDA)~\cite{scaman2017optimal} was developed for solving deterministically smooth and strongly convex distributed optimization problems in networks with a provable optimal linear convergence rate. A new class of decentralized primal-dual methods~\cite{lan2017communication} was also proposed recently in order to improve inter-node communication efficiency for distributed convex optimization problems. To minimize a finite sum of nonconvex functions over a network, the authors in~\cite{hajinezhadzenith} proposed a zeroth-order distributed algorithm (ZENITH) that was globally convergent with a sublinear rate.
%In order to investigate the tradeoff between model accuracy and runtime in distributed deep learning, the authors in~\cite{gupta2015model} introduced a novel learning rate modulation strategy for countering the effect of stale gradients and proposed a new synchronization protocol for the problem, which is able to bound the staleness in gradients.
%Moreover, a stale synchronous parallel model~\cite{ho2013more} to significantly reduce the waiting time for workers was proposed. More recently, elastic averaging SGD (EASGD)~\cite{zhang2015deep} presented a weak consensus among multiple workers by introducing a centralized variable to be optimized in the problem formulation leading local variables to eventually agree on the optimal solution. The gossiping algorithms~\cite{blot2016gossip,jin2016scale} also have been introduced to incorporate with SGD for reducing the training time and the comparison with EASGD showed the encouraging results. CentralVR~\cite{de2016efficient} for reducing the variance and conducting parallel execution with linear convergence rate and Federated Learning~\cite{mcmahan2016communication} were proposed to reduce the communication time cost.
From the perspective of distributed optimization, the proposed algorithms have similarities with the approaches of~\cite{nedic2009distributed,zeng2016nonconvex}. However, we distinguish our work due to the collaborative learning aspect with data parallelization and extension to the stochastic setting and nonconvex objective functions. In~\cite{nedic2009distributed} the authors only considered convex objective functions in a deterministic setting, while the authors in~\cite{zeng2016nonconvex} presented results for non-convex optimization problems in a deterministic setting.

The rest of the paper is organized as follows. While section~\ref{problem_formulation} formulates the distributed, unconstrained stochastic optimization problem, section~\ref{proposed_algorithm} presents the CDSGD algorithm and the Lyapunov stochastic gradient required for analysis presented in section~\ref{main_results}. Validation experiments and performance comparison results are described in section~\ref{experimental_results}. The paper is summarized, concluded in section~\ref{conclusion} along with future research directions. Detailed proofs of analytical results, extensions (e.g., effect of diminishing step size) and additional experiments are included in the supplementary section~\ref{supp}.

%\textbf{Notation}: In this paper, $I$ represents a $N \times N$ identity matrix, and $\mathbf{1}$ denotes a row vector of size $N$ with entries being 1. For a matrix $A$, $A^T$ indicates its transpose while $A_{ij}$ is the $i$th row and $j$th column entry of matrix $A$. The Frobenius norm of $A$ is denoted by $\|A\|_F\triangleq \sqrt{\langle A,A\rangle} = \sqrt{\sum_{i}\sum_{j}A^2_{ij}}$ and when $A$ is a vector it degenerates to the Euclidean norm, which is represented by $\|\cdot\|$.

%%%%%%%%%%%%%%%%%%%%%%%%%%%%%%%%%%%%
%%%%%%%%%%%%%%%%%%%%%%%%%%%%%%%%%%%%
\vspace{-10pt}
\section{Formulation}\label{problem_formulation}
We consider the standard (unconstrained) empirical risk minimization problem typically used in machine learning problems (such as deep learning):
\begin{equation}\label{setup}
\text{min} \, f(x):=\frac{1}{n}\sum_{i=1}^{n}f^i(x) ,
\end{equation}
where $x \in \mathbb{R}^d$ denotes the parameter of interest and $f:\mathbb{R}^d\to\mathbb{R}$ is a given loss function, and $f^i$ is the function value corresponding to a data point $i$.
%Suppose that there exists a training data set, $(a_1,b_1),(a_2,b_2),\dots,(a_n,b_n)$, where $a_i\in \mathbb{R}^d$ and $b_i\in \mathbb{R}$. If the cost function is of the form $f^i(x) = \frac{1}{2}(a_i^Tx-b_i)^2$, then it is linear regression. If the cost function is of the form $f^i(x) = \text{log}(1+\text{exp}(-b_ix^T-a_i))(b_i\in\{-1,1\})$, then it is logistic loss function. Note, that the cost function may include smooth regularization terms~\cite{nitanda2015accelerated}. However, our work extends the above problem setup to distributed setting.
In this paper, we are interested in learning problems where the computational agents exhibit \emph{data parallelism}, i.e., they only have access to their own respective training datasets. However, we assume that the agents can communicate over a static undirected graph $\mathcal{G} = (\mathcal{V},\mathcal{E})$, where $\mathcal{V}$ is a vertex set (with nodes corresponding to agents) and $\mathcal{E}$ is an edge set. With $N$ agents, we have $\mathcal{V} = \{1,2,...,N\}$ and $\mathcal{E} \subseteq \mathcal{V}\times \mathcal{V}$. If $(j,l)\in \mathcal{E}$, then Agent $j$ can communicate with Agent $l$. %As the graph is undirected, if $(j,l)\in \mathcal{E}$ then $(l,j)\in \mathcal{E}$ as well.
The neighborhood of agent $j\in \mathcal{V}$ is defined as: $Nb(j)\triangleq\{l\in \mathcal{V}:(j,l)\in \mathcal{E}\ \text{or}\ j=l\}$. Throughout this paper we assume that the graph $\mathcal{G}$ is \emph{connected}. Let $\mathcal{D}_j,~j=1,\ldots,n$ denote the subset of the training data (comprising $n_j$ samples) corresponding to the $j^{\textrm{th}}$ agents such that $\sum_{j=1}^N n_j=n$. With this setup, we have the following simplification of Eq.~\ref{setup}:
\begin{equation}\label{d_setup_3}
\text{min} f(x):=\frac{1}{n}\sum_{j=1}^{N}\sum_{i\in\mathcal{D}_j}f^i(x) = \frac{N}{n}\sum_{j=1}^{N}\sum_{i\in\mathcal{D}_j}f_j^i(x) ,
\end{equation}
where, $f_j(x) = \frac{1}{N}f(x)$ is the objective function specific to Agent $j$. This formulation enables us to state the optimization problem in a distributed manner, where $f(x) = \sum_{j=1}^N f_j(x)$. \footnote{Note that in our formulation, we are assuming that every agent has the same local objective function while in general distributed optimization problems they can be different.}
Furthermore, the problem \eqref{setup} can be reformulated as
\begin{subequations}
\begin{alignat}{2}
  &\text{min} \, \frac{N}{n}\mathbf{1}^T\F(\x) := \frac{N}{n}\sum_{j=1}^{N}\sum_{i\in \mathcal{D}_j}f^i_j(x^j) \\
  &\text{s.t.~} \, x^j=x^l ~~ \forall (j,l)\in \mathcal{E} ,
\end{alignat}
\end{subequations}
where $\mathbf{x} := (x^1, x^2, \ldots, x^N)^T\in \mathbb{R}^{N\times d}$ and $\F(\x)$ can be written as
\begin{equation}
\F(\x) = \bigg[\sum_{i\in \mathcal{D}_1}f^i_1(x^1),~\sum_{i\in \mathcal{D}_2}f^i_2(x^2),\ldots,~\sum_{i\in \mathcal{D}_N}f^i_N(x^N)\bigg]^T
\end{equation}
Note that with $d\textgreater 1$, the parameter set $\mathbf{x}$ as well as the gradient $\nabla \F(\mathbf{x})$ correspond to matrix variables. However, for simplicity in presenting our analysis, we set $d = 1$ in this paper, which corresponds to the case where $\mathbf{x}$ and $\nabla \F(\mathbf{x})$ are vectors.

\pagebreak
We now introduce several key definitions and assumptions that characterize the objective functions and the agent interaction matrix.
\begin{defn}\label{strong_convexity}
A function $f:\mathbb{R}^d\to \mathbb{R}$ is $H$-strongly convex, if for all $x,y\in \mathbb{R}^d$, we have
$f(y)\geq f(x)+\nabla f(x)^T(y-x)+\frac{H}{2}\|y-x\|^2$.
\end{defn}
%\begin{defn}\label{l_lipschitz}
%A function $f:\mathbb{R}^d\to \mathbb{R}$ is $L$-Lipschitz, if for all $x,y\in \mathbb{R}^d$, we have
%\begin{equation}
%|f(y)-f(x)|\leq L\|y-x\|
%\end{equation}
%\end{defn}
\begin{defn}\label{smooth}
A function $f:\mathbb{R}^d\to \mathbb{R}$ is $\gamma$-smooth if for all $x,y\in \mathbb{R}^d$, we have
$f(y)\leq f(x)+\nabla f(x)^T(y-x)+\frac{\gamma}{2}\|y-x\|^2$.
\end{defn}

As a consequence of Definition~\ref{smooth}, we can conclude that $\nabla f$ is Lipschitz continuous, i.e., $\|\nabla f(y)-\nabla f(x)\|\leq \gamma\|y-x\|$~\cite{NO16}.

\begin{defn}\label{coercivity}
A function $c$ is said to be coercive if it satisfies:
$c(x)\to \infty\;when \|x\|\to \infty.$
\end{defn}

\begin{assumption}\label{assump_objective}
The objective functions $f_j:\mathbb{R}^d\to\mathbb{R}$ are assumed to satisfy the following conditions: a) Each $f_j$ is $\gamma_j$-smooth; b) each $f_j$ is proper (not everywhere infinite) and coercive; and c) each $f_j$ is $L_j$-Lipschitz continuous, i.e., $|f_j(y)-f_j(x)|<L_j\|y-x\|~\forall x,y\in\mathbb{R}^d$.
\end{assumption}
As a consequence of Assumption~\ref{assump_objective}, we can conclude that $\sum_{j=1}^{N}f_j(x^j)$ possesses Lipschitz continuous gradient with parameter $\gamma_m:=\text{max}_j\gamma_{j}$. Similarly, each $f_j$ is strongly convex with $H_{j}$ such that $\sum_{j=1}^{N}f_j(x^j)$ is strongly convex with $H_m = \text{min}_jH_{j}$.
%\color{blue} Moreover, another consequence that can be achieved is that the gradient of $\sum_{j=1}^{N}f_j(x_j)$ is bounded above by $L_m = \text{max}_jL_j$. \color{black}

Regarding the communication network, we use $\Pi$ to denote the agent interaction matrix, where the element $\pi_{jl}$ signifies the link weight between agents $j$ and $l$.

\begin{assumption}\label{assump_1}
a) If $(j,l)\notin \mathcal{E}$, then $\pi_{jl} = 0$; b) $\mathbf{1}^T\Pi = \mathbf{1}^T, \Pi\mathbf{1} = \mathbf{1}$; c) $\text{null}\{I-\Pi\} = \text{span}\{\mathbf{1}\}$; and d) $I\succeq \Pi\succ 0$.
\end{assumption}
The main outcome of Assumption~\ref{assump_1} is that the probability transition matrix is doubly stochastic and that we have $\lambda_1(\Pi) = 1\textgreater\lambda_2(\Pi)\geq\dots \geq\lambda_N(\Pi)\textgreater 0$, where $\lambda_z(\Pi)$ denotes the $z$-th largest eigenvalue of $\Pi$. %\red{Assumption~\ref{assump_1}(d) may be relaxed in the future work.}
%\textbf{Equivalence}: Either centralized or distributed problem setup should be reflected to solve the minimization problem such that they are equivalent. Next we now show the equivalence of these two problem setups. As the cost function corresponding to each training sample in the centralized manner is $f^i(x)$, then in the distributed manner for $N$ agents they have the function values $f^i_j(x)=\frac{1}{N}f^i(x)$ corresponding to each training sample. With such a relation, it can be observed that $\sum_{j=1}^{N}\frac{N}{n}\sum_{i=1}^{\frac{n}{N}}\frac{1}{N}f^i(x) = \sum_{j=1}^{N}\frac{1}{n}\sum_{i=1}^{\frac{n}{N}}f^i(x)$, which implies that $\frac{1}{n}N\sum_{i=1}^{\frac{n}{N}}f^i(x) = \frac{1}{n}\sum_{i=1}^{n}f^i(x)$. The last equality follows from that the intersection of two data subsets is empty set.

\vspace{-10pt}
\section{Proposed Algorithm}\label{proposed_algorithm}
\vspace{-10pt}
\subsection{Consensus Distributed SGD}\vspace{-8pt}
For solving stochastic optimization problems, SGD and its variants have been commonly used to centralized and distributed problem formulations. Therefore, the following algorithm is proposed based on SGD and the concept of consensus to solve the problem laid out in Eq.~\ref{d_setup_3},
\begin{equation}\label{update_law}
%x^j_{k+1} = \sum_{l\in Nb(j)}(\pi)_{jl}x^l_k-\alpha_k\frac{1}{b'}\sum_{q'\in \mathcal{B'}}\nabla f^{q'}_j( x^j_k)
x^j_{k+1} = \sum_{l\in Nb(j)}\pi_{jl}x^l_k-\alpha g_j(x^j_k)
\end{equation}
where $Nb(j)$ indicates the neighborhood of agent $j$, $\alpha$ is the step size, $g_j(x^j_k)$ is stochastic gradient of $f_j$ at $x^j_k$.
%which corresponds to a minibatch of sampled data points at the $k^{th}$ epoch. More formally, $g_j(x^j_k)=\frac{1}{b'}\sum_{q'\in \mathcal{D'}}\nabla f^{q'}_j(x^j_k)$,
%\begin{equation}
%g_j(x^j_k)=
%\begin{cases}
%  \nabla f^i_j(x^j_k) \\
%  \frac{1}{b'}\sum_{q'\in \mathcal{D'}}\nabla f^{q'}_j(x^j_k).
%\end{cases}
%\end{equation}
%where $b'$ is the size of the minibatch $\mathcal{D'}$ randomly selected from the data subset $\mathcal{D}_j$.
While the pseudo-code of CDSGD is shown below in Algorithm~\ref{CDSGD}, momentum versions of CDSGD based on Polyak momentum~\cite{polyak1964some} and Nesterov momentum~\cite{nesterov2013introductory} are also presented in the supplementary section~\ref{supp}. Note, mini-batch implementations of these algorithms are straightforward, hence, are not discussed here in detail.

\begin{algorithm}[H]\label{CDSGD}
    \caption{CDSGD}
    \SetKwInOut{Input}{Input}
    \SetKwInOut{Output}{Output}

    \Input{~$m$, $\alpha$, $N$}
    \textbf{Initialize:}~\text{$x^j_0, (j=1,2,\dots, N$)}\\
    \text{Distribute the training dataset to $N$ agents.}\\
    \For{each agent}{
    \For{$k=0:m$}
      {
        \text{Randomly shuffle the corresponding data subset $\mathcal{D}_j$}\\
        $w_{k+1}^j = \sum_{l\in Nb(j)}\pi_{jl}x^l_{k}$\\
        $x^j_{k+1} = w_{k+1}^j -\alpha g_j(x^j_{k})$
      }
      }
\end{algorithm}

\vspace{-10pt}
\subsection{Tools for convergence analysis}\vspace{-8pt}
We now analyze the convergence properties of the iterates $\{x^j_k\}$ generated by Algorithm~\ref{CDSGD}. The following section summarizes some key intermediate concepts required to establish our main results. % (with several straightforward  omitted due to lack of space.)

First, we construct an appropriate \emph{Lyapunov} function that will enable us to establish convergence. Observe that the update law in Alg. \ref{CDSGD} can be expressed as:
\begin{equation}\label{vector_form}
\x_{k+1} = \Pi\x_k -\alpha\mathbf{g}(\x_k) ,
\end{equation}
where
%\[
%\mathbf{g}(\x_k) = \bigg[\nabla f^{i_1}_1(x_k^1)\;\nabla f^{i_2}_2(x_k^2)\dots \nabla f^{i_N}_N(x_k^N)\bigg]^T
%\]
%or
%\[
%\mathbf{g}(\x_k) = \bigg[\frac{1}{b'_1}\sum_{q'\in \mathcal{D}'_1}\nabla f^{q'}_1,~\frac{1}{b'_2}\sum_{q'\in \mathcal{D}'_2}\nabla f^{q'}_2,~\dots, \frac{1}{b'_N}\sum_{q'\in \mathcal{D}'_N}\nabla f^{q'}_N\bigg]^T
%\]
\[
\mathbf{g}(\x_k) = [g_1(x^1_k) g_2(x^2_k)...g_N(x^N_k)]^T
\]

%and where $b'_1, b'_2,\dots, b'_N$ are the sizes of the corresponding mini-batches $\mathcal{D}'_1, \mathcal{D}'_2, \dots, \mathcal{D}'_N$ of samples that are randomly selected from the data subsets $\mathcal{D}_1, \mathcal{D}_2, \dots, \mathcal{D}_N$, respectively.

Denoting $\w_k = \Pi\x_k$, the update law can be re-written as $\x_{k+1} =  \w_k-\alpha\mathbf{g}(\x_k)$. Moreover, $\x_{k+1} = \x_k-\x_k+\w_k-\alpha\mathbf{g}(\x_k)$. Rearranging the last equality yields the following relation:
\begin{equation}\label{reformulation}
\x_{k+1} = \x_k - \alpha(\mathbf{g}(\x_k)+\alpha^{-1}(\x_k-\w_k))=\x_k - \alpha{(\mathbf{g}(\x_k)+\alpha^{-1}(I-\Pi)\x_k)}%_{\text{Stochastic Lyapunov Gradient}}
\end{equation}
where the last term in Eq.\ \ref{reformulation} is the \emph{Stochastic Lyapunov Gradient}. From Eq.~\ref{reformulation}, we observe that the ``effective" gradient step is given by $\mathbf{g}(\x_k)+\alpha^{-1}(I-\Pi)\x_k$.  Rewriting $\nabla \mathcal{J}^i(\x_k) = \mathbf{g}(\x_k)+\alpha^{-1}(I-\Pi)\x_k$, the updates of CDSGD can be expressed as:
\begin{equation}\label{CDSGD_fixed stepsize}
\x_{k+1} = \x_k-\alpha\nabla \mathcal{J}^i(\x_k).
\end{equation}

The above expression naturally motivates the following Lyapunov function candidate:
\begin{equation}\label{lyapunov_func}
V(\x,\alpha) := \frac{N}{n}\mathbf{1}^T\F(\x) + \frac{1}{2\alpha}\|\x\|_{I-\Pi}^2
\end{equation}
%where $\alpha$ here indicates one parameter associated with the Lyapunov function, $\|\x\|_{I-\Pi}$ is the Frobenius norm in this context. It is noted that for fixed step size, we use $\tilde{\alpha}$ to replace $\alpha$ while for diminishing step size, it is still $\alpha_k$.
where $\|\cdot\|_{I-\Pi}$ denotes the norm with respect to the PSD matrix $I - \Pi$.
%Such Lyapunov function has the similar form as proposed in~\cite{zeng2016nonconvex} in which the authors only considered the nonconvex objective functions in a deterministic setting. However, in this work we employ the algorithm in a stochastic setting to strongly convex objective functions and to nonconvex objective functions with fixed and diminishing step sizes. We introduce several key definitions and assumptions that characterize the objective functions and probability transition matrix $\Pi$.
Since $\sum_{j=1}^{N}f_j(x^j)$ has a $\gamma_m$-Lipschitz continuous gradient, $\nabla V(\x)$ also has a Lipschitz continuous gradient with parameter:
$$\hat{\gamma}:=\gamma_m+\alpha^{-1}\lambda_{\text{max}}(I-\Pi) = \gamma_m+\alpha^{-1}(1-\lambda_N(\Pi)).$$
Similarly, as $\sum_{j=1}^{N}f_j(x^j)$ is $H_m$-strongly convex, then $V(\x)$ is strongly convex with parameter:
$$\hat{H}:= H_m+(2\alpha)^{-1}\lambda_{\text{min}}(I-\Pi) = H_m+(2\alpha)^{-1}(1-\lambda_2(\Pi)).$$

Based on Definition~\ref{strong_convexity}, $V$ has a unique minimizer, denoted by $\x^*$ with $V^* = V(\x^*)$. Correspondingly, using strong convexity of $V$, we can obtain the relation:
\begin{equation}\label{strong_convexity_formula}
2\hat{H}(V(\x)-V^*)\leq \|\nabla V(\x)\|^2 \;\text{for all $\x\in\mathbb{R}^N$}.
\end{equation}
From strong convexity and the Lipschitz continuous property of $\nabla f_j$, the constants $H_m$ and $\gamma_m$ further satisfy $H_m\leq \gamma_m$ and hence, $\hat{H}\leq \hat{\gamma}$.

Next, we introduce two key lemmas that will help establish our main theoretical guarantees. Due to space limitations, all proofs are deferred to the supplementary material in Section~\ref{supp}.
\begin{lem}\label{lemma_1}
Under Assumptions~\ref{assump_objective} and~\ref{assump_1}, the iterates of CDSGD satisfy $\forall k\in \mathbb{N}$:
\begin{equation}
\mathbb{E}[V(\x_{k+1})]-V(\x_k) \leq -\alpha\nabla V(\x_k)^T\mathbb{E}[\nabla \mathcal{J}^i(\x_k)]+\frac{\hat{\gamma}}{2}\alpha^2\mathbb{E}[\|\nabla \mathcal{J}^i(\x_k)\|^2]
\end{equation}
\end{lem}

At a high level, since $\mathbb{E}[\nabla \mathcal{J}^i(\x_k)]$ is the unbiased estimate of $\nabla V(\x_k)$, using the updates $\nabla \mathcal{J}^i(\x_k)$ will lead to sufficient decrease in the Lyapunov function. However, unbiasedness is not enough, and we also need to control higher order moments of $\nabla \mathcal{J}^i(\x_k)$ to ensure convergence. Specifically, we consider the variance of $\nabla \mathcal{J}^i(\x_k)$:
\begin{equation}\label{variance}
Var[\nabla \mathcal{J}^i(\x_k)]:=\E[\|\nabla \mathcal{J}^i(\x_k)\|^2]-\|\E[\nabla \mathcal{J}^i(\x_k)]\|^2
\end{equation}
To bound the variance of $\nabla \mathcal{J}^i(\x_k)$, we use a standard assumption presented in~\cite{bottou2016optimization} in the context of (centralized) deep learning. %The assumption has been justified for SGD and in this context we extend it to the distributed SGD.
Such an assumption aims at providing an upper bound for the ``gradient noise" caused by the randomness in the minibatch selection at each iteration.
\begin{assumption}\label{assumption_2}
a) There exist scalars $\zeta_2\geq\zeta_1\textgreater 0$ such that $\nabla V(\x_k)^T\E[\nabla \mathcal{J}^i(\x_k)]\geq \zeta_1\|\nabla V(\x_k)\|^2$ and $\|\E[\nabla \mathcal{J}^i(\x_k)]\|\leq \zeta_2\|\nabla V(\x_k)\|$ for all $k\in \N$; b) There exist scalars $Q\geq 0$ and $Q_V\geq 0$ such that $Var[\nabla \mathcal{J}^i(\x_k)]\leq Q+Q_V\|\nabla V(\x_k)\|^2$ for all $k\in\N$.
\end{assumption}
\begin{rem}\label{rem1}
While Assumption~\ref{assumption_2}(a) guarantees the sufficient descent of $V$ in the direction of $-\nabla \mathcal{J}^i(\x_k)$, Assumption~\ref{assumption_2}(b) states that the variance of $\nabla \mathcal{J}^i(\x_k)$ is bounded above by the second moment of $\nabla V(\x_k)$. The constant $Q$ can be considered to represent the second moment of the ``gradient noise" in $\nabla \mathcal{J}^i(\x_k)$. Therefore, the second moment of $\nabla \mathcal{J}^i(\x_k)$ can be bounded above as $\E[\|\nabla \mathcal{J}^i(\x_k)\|^2]\leq Q+Q_m\|\nabla V(\x_k)\|^2$, where $Q_m := Q_V+\zeta_2^2\geq \zeta_1^2\textgreater 0$.
\end{rem}
\begin{lem}\label{lemma2}
Under Assumptions~\ref{assump_objective},~\ref{assump_1}, and~\ref{assumption_2}, the iterates of CDSGD satisfy~$\forall k\in \mathbb{N}$:
\begin{equation}\label{lem2}
\E[V(\x_{k+1})]-V(\x_k)\leq -(\zeta_1-\frac{\hat{\gamma}}{2}\alpha Q_m)\alpha\|\nabla V(\x_k)\|^2+\frac{\hat{\gamma}}{2}\alpha^2Q \, .
\end{equation}
\end{lem}
%\begin{proof}
%Recalling Lemma~\ref{lemma_1} and using Assumption~\ref{assumption_2} and Remark~\ref{rem1}, we have
%\begin{equation}
%\begin{aligned}
%\E[V(\x_{k+1})]-V(\x_k)&\leq -\zeta_1\alpha\|\nabla V(\x_k)\|^2+\frac{\hat{\gamma}}{2}\alpha\E[\|\nabla \mathcal{J}^i(\x_k)\|^2]\\
%                       &\leq -\zeta_1\alpha\|\nabla V(\x_k)\|^2+\frac{\hat{\gamma}}{2}\alpha(Q+Q_m\|\nabla V(\x_k)\|^2)\\
%                       &=-\zeta_1\alpha\|\nabla V(\x_k)\|^2+\frac{\hat{\gamma}}{2}\alpha^2Q+\frac{\hat{\gamma}}{2}\alpha^2\\
%                       &Q_m\|\nabla V(\x_k)\|^2\\
%                       &=-(\zeta_1-\frac{\hat{\gamma}}{2}\alpha Q_m)\alpha\|\nabla V(\x_k)\|^2+\frac{\hat{\gamma}}{2}\alpha^2Q
%\end{aligned}
%\end{equation}
%which completes the proof.
%\end{proof}
%The first term on the right hand side of Lemma~\ref{lemma2} can be strictly negative when $\alpha$ is sufficiently small and implies the decrease associated with the objective function by a magnitude proportional to the second moment of gradient, i.e., $\|\nabla V(\x_k)\|^2$. Nevertheless, the second term may be large enough to make the objective function value increase. Design of the optimization methods to enable the sufficient decrease in the objective function value is significantly critical.
In Lemma~\ref{lemma2}, the first term is strictly negative if the step size satisfies the following necessary condition: \color{black}
\begin{equation}\label{step_size_condition}
0\textless \alpha\leq\frac{2\zeta_1}{\hat{\gamma}Q_m}
\end{equation}
However, in latter analysis, when such a condition is substituted into the convergence analysis, it may produce a larger upper bound. For obtaining a tight upper bound, we impose a sufficient condition for the rest of analysis as follows:
\begin{equation}\label{step_size_condition_2}
0\textless \alpha\leq\frac{\zeta_1}{\hat{\gamma}Q_m}
\end{equation}
As $\hat{\gamma}$ is a function of $\alpha$, the above inequality can be rewritten as $0\textless \alpha\leq\frac{\zeta_1-(1-\lambda_N(\Pi))Q_m}{\gamma_mQ_m}$.

\vspace{-10pt}
\section{Main Results}\label{main_results}\vspace{-10pt}
We now present our main theoretical results establishing the convergence of CDSGD. First, we show that for most generic loss functions (whether convex or not), CDSGD achieves \emph{consensus} across different agents in the graph, provided the step size (which is fixed across iterations) does not exceed a natural upper bound.
\begin{proposition}{(Consensus with fixed step size)}\label{prop1}
Under Assumptions~\ref{assump_objective} and~\ref{assump_1}, the iterates of CDSGD (Algorithm~\ref{CDSGD}) satisfy~$\forall k\in \mathbb{N}$:
\begin{equation}
\E[\|x^j_k - s_k\|]\leq \frac{\alpha L}{1-\lambda_2(\Pi)}
\end{equation}
where $\alpha$ satisfies $0\textless \alpha\leq\frac{\zeta_1-(1-\lambda_N(\Pi))Q_m}{\gamma_mQ_m}$ and $L$ is an upper bound of $\mathbb{E} [\|\mathbf{g}(\x_k)\|], \forall k\in \mathbb{N}$ (defined properly and discussed in Lemma~\ref{lemma6} in the supplementary section~\ref{supp}) and $s_k = \frac{1}{N}\sum_{j=1}^{N}x_k^j$ represents the average parameter estimate.
\end{proposition}
The proof of this proposition can be adapted from~\cite[Lemma 1]{yuan2013convergence}.
%formally, we show it in Lemma~\ref{lemma6} which is provided in the supplementary material.

Next, we show that for strongly convex loss functions, CDSGD converges linearly to a neighborhood of the global optimum.
\begin{thm}\label{theorem1}{(Convergence of CDSGD with fixed step size, strongly convex case)}
Under Assumptions~\ref{assump_objective},~\ref{assump_1} and~\ref{assumption_2}, the iterates of CDSGD satisfy the following inequality $\forall k\in \mathbb{N}$:
\begin{equation}
\begin{aligned}
\E[V(\x_k)-V^*]&\leq(1-\alpha \hat{H}\zeta_1)^{k-1}(V(\x_1)-V^*)+\frac{\alpha^2\hat{\gamma}Q}{2}\sum_{l=0}^{k-1}(1-\alpha\hat{H}\zeta_1)^l\\
               &=(1-(\alpha H_m+1-\lambda_2(\Pi))\zeta_1)^{k-1}(V(\x_1)-V^*)\\
               &+\frac{(\alpha^2\gamma_m+\alpha(1-\lambda_N(\Pi)))Q}{2}\sum_{l=0}^{k-1}(1-(\alpha H_m+1-\lambda_2(\Pi))\zeta_1)^{l}
\end{aligned}
\end{equation}
when the step size satisfies $0\textless \alpha\leq\frac{\zeta_1-(1-\lambda_N(\Pi))Q_m}{\gamma_mQ_m}$.
\end{thm}
%\begin{proof}
%Recalling Lemma~\ref{lemma2} and using Eq.~\ref{strong_convexity_formula} yield that
%\begin{equation}
%\begin{aligned}
%\E[V(\x_{k+1})]-V(\x_k)&\leq -(\zeta_1-\frac{\hat{\gamma}}{2}\alpha Q_m)\alpha\|\nabla V(\x_k)\|^2+\frac{\hat{\gamma}}{2}\alpha^2Q\\
%                       &\leq -\frac{1}{2}\alpha \zeta_1\|\nabla(\x_k)\|^2+\frac{\alpha^2\hat{\gamma}Q}{2}\\
%                       &\leq -\alpha\zeta_1\hat{H}(V(\x_k)-V^*)+\frac{\alpha^2\hat{\gamma}Q}{2}
%\end{aligned}
%\end{equation}
%The second inequality follows from that $\alpha\leq \frac{\zeta_1}{\hat{\gamma}Q_m}$, which is implied by that $\alpha\leq\frac{\zeta_1-(1-\lambda_N(\Pi))Q_m}{\gamma_mQ_m}$. The expectation taken in the above inequalities is only related to $\x_{k+1}$. Hence, recursively taking the expectation and subtracting $V^*$ from both sides require it to satisfy the following
%\begin{equation}\label{thm1}
%\E[V(\x_{k+1})-V^*]\leq (1-\alpha\hat{H}\zeta_1)\E[V(\x_k)-V^*]+\frac{\alpha^2\hat{\gamma}Q}{2}
%\end{equation}
%As $0\textless \alpha\hat{H}\zeta_1\leq\frac{\hat{H}\zeta_1^2}{\hat{\gamma}Q_m}\leq \frac{\hat{H}\zeta_1^2}{\hat{\gamma}\zeta_1^2}=\frac{\hat{H}}{\hat{\gamma}}\leq 1$, the conclusion follows by applying Eq.~\ref{thm1} recursively through iteration $k\in \N$.
%\end{proof}
%\begin{rem}
A detailed proof is presented in the supplementary section~\ref{supp}. We observe from Theorem~\ref{theorem1} that the sequence of Lyapunov function values $\{V(\x_k)\}$ converges \emph{linearly} to a \emph{neighborhood} of the optimal value, i.e., $\lim_{k\to \infty}\E[V(\x_k)-V^*]\leq \frac{\alpha\hat{\gamma}Q}{2\hat{H}\zeta_1} = \frac{(\alpha\gamma_m+1-\lambda_N(\Pi))Q}{2(H_m+\alpha^{-1}(1-\lambda_2(\Pi))\zeta_1}$. We also observe that the term on the right hand side decreases with the spectral gap of the agent interaction matrix $\Pi$, i.e., $1-\lambda_2(\Pi)$, which suggests an interesting relation between convergence and topology of the graph. Moreover, we observe that the upper bound is proportional to the step size parameter $\alpha$, and smaller step sizes lead to smaller radii of convergence. (However, choosing a very small step-size may negatively affect the convergence \emph{rate} of the algorithm). Finally, if the gradient in this context is \emph{not} stochastic (i.e., the parameter $Q = 0$), then linear convergence to the optimal value is achieved, which matches known rates of convergence with (centralized) gradient descent under strong convexity and smoothness assumptions.
%\end{rem}

\begin{rem}
Since $\E[\frac{N}{n}\mathbf{1}^T\F(\x_k)]\leq \E[V(\x_k)]$ and $\frac{N}{n}\mathbf{1}^T\F(\x^*) = V^*$, the sequence of objective function values are themselves upper bounded as follows:
%\begin{align*}
$\E[\frac{N}{n}\mathbf{1}^T\F(\x_k)-\frac{N}{n}\mathbf{1}^T\F(\x^*)] \leq \E[V(\x_k)-V^*]$.
%&\leq(1-\alpha \hat{H}\zeta_1)^{k-1}(V(\x_1)-V^*)+\frac{\alpha^2\hat{\gamma}Q}{2}\sum_{l=0}^{k-1}(1-\alpha\hat{H}\zeta_1)^l.
%\end{align*}
Therefore, using Theorem~\ref{theorem1} we can establish analogous convergence rates in terms of the true objective function values $\{\frac{N}{n}\mathbf{1}^T\F(\x_k)\}$ as well.
\end{rem}

The above convergence result for CDSGD is limited to the case when the objective functions are strongly convex. However, most practical deep learning systems (such as convolutional neural network learning) involve optimizing over highly \emph{non-convex} objective functions, which are much harder to analyze.
%Solving a nonconvex optimization problem is much more challenging than doing that for a convex case as the global optimal is not guaranteed.
Nevertheless, we show that even under such situations, CDSGD exhibits a (weaker) notion of convergence.
\begin{thm}\label{thm2}{(Convergence of CDSGD with fixed step size, nonconvex case)}
Under Assumptions~\ref{assump_objective},~\ref{assump_1}, and~\ref{assumption_2}, the iterates of CDSGD satisfy~$\forall m\in \mathbb{N}$:
\begin{equation}
\label{part1_thm2}
\begin{aligned}
\E[\sum_{k=1}^{m}\|\nabla V(\x_k)\|^2]&\leq \frac{\hat{\gamma}m\alpha Q}{\zeta_1}+\frac{2(V(\x_1)-V_{\text{inf}})}{\zeta_1\alpha}\\
                                      &=\frac{(\gamma_m\alpha+1-\lambda_N(\Pi))mQ}{\zeta_1}+\frac{2(V(\x_1)-V_{\text{inf}})}{\zeta_1\alpha} .
\end{aligned}
\end{equation}
when the step size satisfies
$0 < \alpha\leq\frac{\zeta_1-(1-\lambda_N(\Pi))Q_m}{\gamma_mQ_m}$.
%and thus
%\begin{equation}
%\frac{1}{m}\E[\sum_{k=1}^{m}\|\nabla V(\x_k)\|^2]\leq \frac{(\gamma_m\alpha+1-\lambda_N(\Pi))Q}{\zeta_1}+\frac{2(V(\x_1)-V_{\text{inf}})}{m\zeta_1\alpha}
%\end{equation}
\end{thm}
%\begin{proof}
%Recalling Lemma~\ref{lemma2}, and also taking the expectation lead to the following relation,
%\begin{equation}\label{nonconvex_descent}
%\E[V(\x_{k+1})]-\E[V(\x_k)]\leq -(\zeta_1-\frac{\hat{\gamma}\alpha Q_m}{2})\alpha\E[\|\nabla V(\x_k)\|^2]+\frac{\hat{\gamma}\alpha^2Q}{2}
%\end{equation}
%As the step size satisfies that $\alpha\leq\frac{\zeta_1}{\hat{\gamma}Q_m}$, it results in
%\begin{equation}
%\E[V(\x_{k+1})]-\E[V(\x_k)]\leq  -\frac{\zeta_1\alpha}{2}\E[\|\nabla V(\x_k)\|^2]+\frac{\alpha^2\hat{\gamma}Q}{2}
%\end{equation}
%Applying the above inequality from 1 to $m$ and summing them up can give the following relation
%\begin{equation}
%V_{\text{inf}} - V(\x_1)\leq \E[V(\x_{k+1})]-V(\x_1)\leq -\frac{\zeta_1\alpha}{2}\sum_{k=1}^{m}\E[\|\nabla V(\x_k)\|^2]+\frac{m\alpha^2\hat{\gamma}Q}{2}
%\end{equation}
%The last inequality follows from the Assumption~\ref{assumption_2}. Rearrangement of the above inequality and substituting $\hat{\gamma} = \gamma_m+\alpha^{-1}(1-\lambda_N(\Pi))$ into it yield the first part of results. It is immediately that when both sides are divided by $m$ the second part of results holds.
%\end{proof}
\begin{rem}
Theorem~\ref{thm2} states that when in the absence of ``gradient noise" (i.e., when $Q = 0$), the quantity $\E[\sum_{k=1}^{m}\|\nabla V(\x_k)\|^2]$ remains finite. Therefore, necessarily $\{\|\nabla V(\x_k)\|\} \to 0$ and the estimates approach a stationary point. On the other hand, if the gradient calculations are stochastic, then a similar claim cannot be made. %Therefore gradient noise inhibits the further asymptotic convergence to 0 for the running average of the second moment of gradients in a distributed setting.
However, for this case we have the upper bound $\lim_{m\to\infty}\E[\frac{1}{m}\sum_{k=1}^{m}\|\nabla V(\x_k)\|^2]\leq \frac{(\gamma_m\alpha+1-\lambda_N(\Pi))Q}{\zeta_1}$. This tells us that while we cannot guarantee convergence in terms of sequence of objective function values, we can still assert that the average of the second moment of gradients is strictly bounded from above even for the case of nonconvex objective functions.

Moreover, the upper bound cannot be solely controlled via the step-size parameter $\alpha$ (which is different from what is implied in the strongly convex case by Theorem~\ref{theorem1}). In general, the upper bound becomes tighter as $\lambda_N(\Pi)$ increases; however, an increase in $\lambda_N(\Pi)$ may result in a commensurate increase in $\lambda_2(\Pi)$, leading to worse connectivity in the graph and adversely affecting consensus among agents. Again, our upper bounds are reflective of interesting tradeoffs between consensus and convergence in the gradients, and their dependence on graph topology.
\end{rem}

The above results are for \emph{fixed} step size $\alpha$, and we can prove complementary results for CDSGD even for the (more prevalent) case of \emph{diminishing} step size $\alpha_k$. These are presented in the supplementary material due to space constraints.

\vspace{-10pt}
\pagebreak
\section{Experimental Results}\label{experimental_results}\vspace{-10pt}
This section presents the experimental results using the benchmark image recognition dataset, CIFAR-10. We use a deep convolutional nerual network (CNN) model (with 2 convolutional layers with 32 filters each followed by a max pooling layer, then 2 more convolutional layers with 64 filters each followed by another max pooling layer and a dense layer with 512 units, ReLU activation is used in convolutional layers) to validate the proposed algorithm. We use a fully connected topology with 5 agents and uniform agent interaction matrix except mentioned otherwise. A mini-batch size of 128 and a fixed step size of 0.01 are used in these experiments. The experiments are performed using Keras and TensorFlow~\cite{chollet2015keras,abadi2016tensorflow} and the codes will be made publicly available soon. While we included the training and validation accuracy plots for the different case studies here, the corresponding training loss plots, results with other becnmark datasets such as MNIST and CIFAR-100 and decaying as well as different fixed step sizes are presented in the supplementary section~\ref{supp}.

\vspace{-10pt}
\subsection{Performance comparison with benchmark methods}
\begin{figure}
\centering
\subfigure[]{\includegraphics[width=0.48\textwidth]{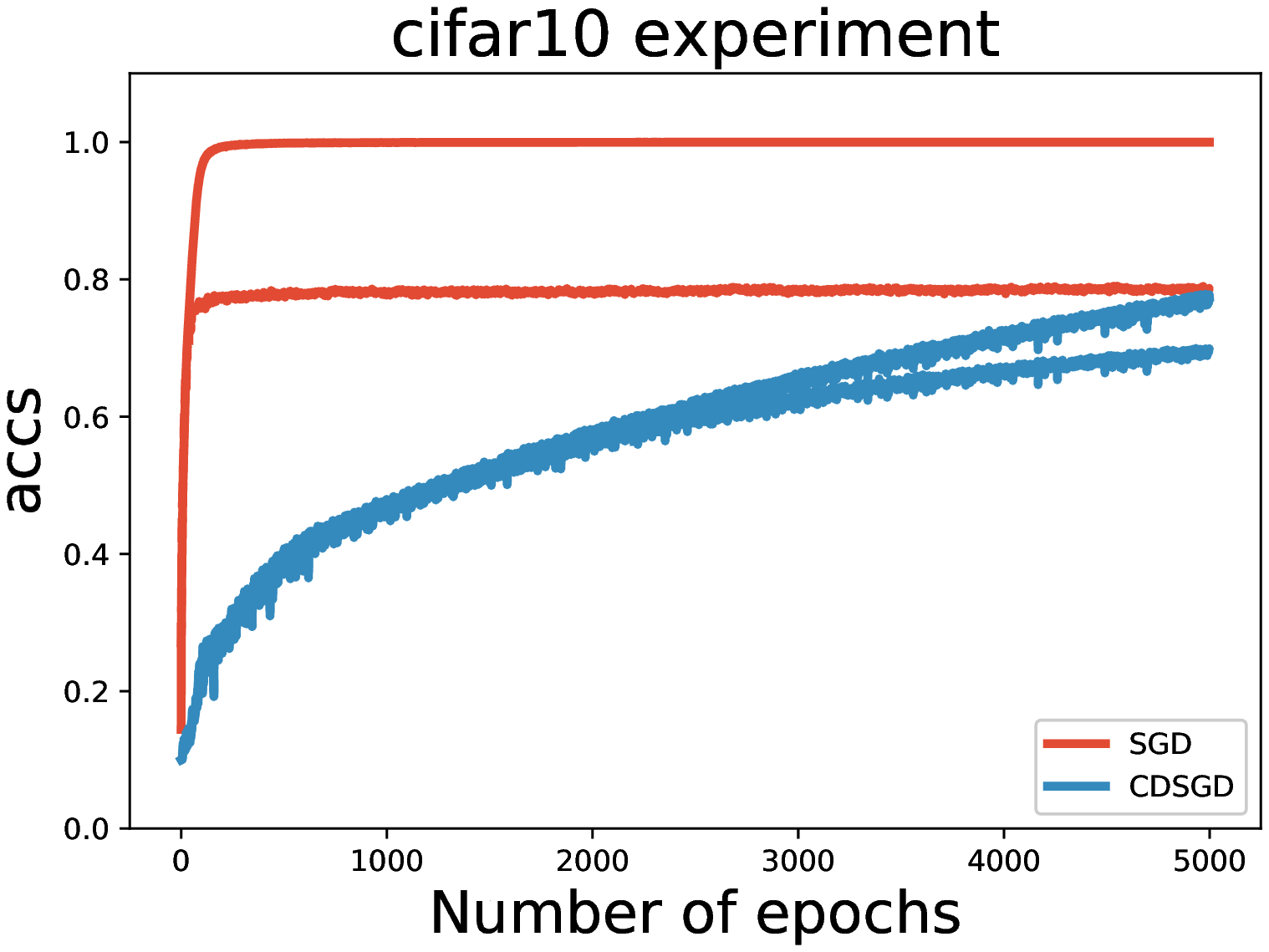}}
\subfigure[]{\includegraphics[width=0.48\textwidth]{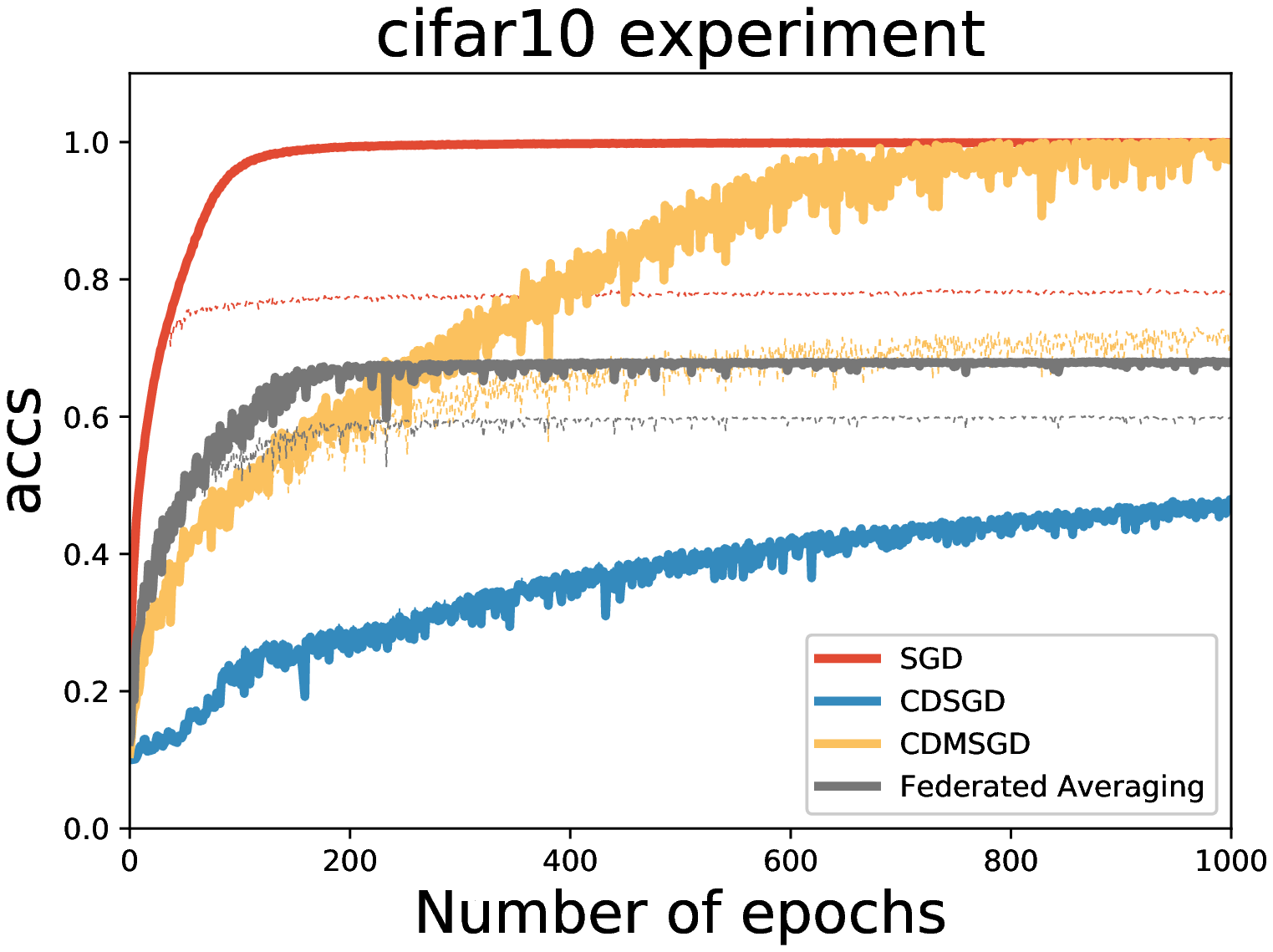}}\vspace{-10pt}
\caption{\textit{Average training (solid lines) and validation (dash lines) accuracy for (a) comparison of CDSGD with centralized SGD and (b) CDMSGD with Federated average method}}\label{Figure1}\vspace{-15pt}
\end{figure}

\vspace{-8pt}
We begin with comparing the accuracy of CDSGD with that of the centralized SGD algorithm as shown in Fig.~\ref{Figure1}(a). While the CDSGD convergence rate is significantly slower compared to SGD as expected, it is observed that CDSGD can eventually achieve high accuracy, comparable with centralized SGD. However, another interesting observation is that the generalization gap (the difference between training and validation accuracy as defined in~\cite{ZBHR16}) for the proposed CDSGD algorithm is significantly smaller than that of SGD which is an useful property.
%different algorithms in terms of accuracy. A significant implication can be made that the generalization gap (the difference between training accuracy and test accuracy) for the proposed CDSGD is smallest while SGD has the largest gap. Moreover, CDMSGD is able to perform better than FedAvg. Thus, based on Fig.~\ref{Figure1}, the proposed CDSGD is a robust method compared to SGD and FedAvg and the momentum term can be chosen appropriately to not only improve the accuracy but to keep the small generalization gap.
%This section presents
%the experimental results using two benchmark datasets, namely, CIFAR-10 and CIFAR-100 to validate the proposed algorithm. We use convolutional nerual networks (CNN) model to train and test the datasets. Except results of dependence on number of agents, we fix the number of agents 4 for experiments. The constant step size is 0.01 in this context while the diminishing step size is $\frac{1}{\sqrt{1+ak}}$, where $a\textgreater 0$ and $k$ is the number of epochs. The communication period $\tau$ is 1 unless specified under some certain situations.
We also compare both CDSGD and CDMSGD with the Federated averaging SGD (FedAvg) algorithm which also performs data parallelization (see Fig.~\ref{Figure1}(b)). For the sake of comparison, we use same number of agents and choose $E=1$ and $C=1$ as the hyperparameters in the FedAvg algorithm as it is close to a fully connected topology scenario as considered in the CDSGD and CDMSGD experiments. As CDSGD is significantly slow, we mainly compare the CDMSGD with FedAvg which have similar convergence rates (CDMSGD being slightly slower). The main observation is that CDMSGD performs better than FedAvg at the steady state and can achieve centralized SGD level performance. It is important to note that FedAvg does not perform decentralized computation. Essentially it runs a brute force parameter averaging on a central parameter server at every epoch (i.e., consensus at every epoch) and then broadcasts the updated parameters to the agents. Hence, it tends to be slightly faster than CDMSGD which uses a truly decentralized computation over a network.
%Figure~\ref{Figure1} shows the loss and accuracy with respect to the number of epochs for SGD, CDSGD, FedAvg and CDMSGD. The solid curve means training and the dash curve indicates testing. From the loss results, it can be observed that SGD has the sublinear convergence rate for training and testing and is dominated among the discussed methods for the training. While for the testing, SGD performs poorly after around 70 epochs. For the proposed CDSGD/CDMSGD, they show the linear convergence rate for training and CDMSGD has the better testing performance than SGD. Although FedAvg shows a quick decrease on objective function at the beginning, along with epochs, it shows the worst convergence rate. We also compare different algorithms in terms of accuracy. A significant implication can be made that the generalization gap (the difference between training accuracy and test accuracy) for the proposed CDSGD is smallest while SGD has the largest gap. Moreover, CDMSGD is able to perform better than FedAvg. Thus, based on Fig.~\ref{Figure1}, the proposed CDSGD is a robust method compared to SGD and FedAvg and the momentum term can be chosen appropriately to not only improve the accuracy but to keep the small generalization gap.

\vspace{-10pt}
\subsection{Effect of network size and topology}
\begin{figure}
\centering
\subfigure[]{\includegraphics[width=0.48\textwidth]{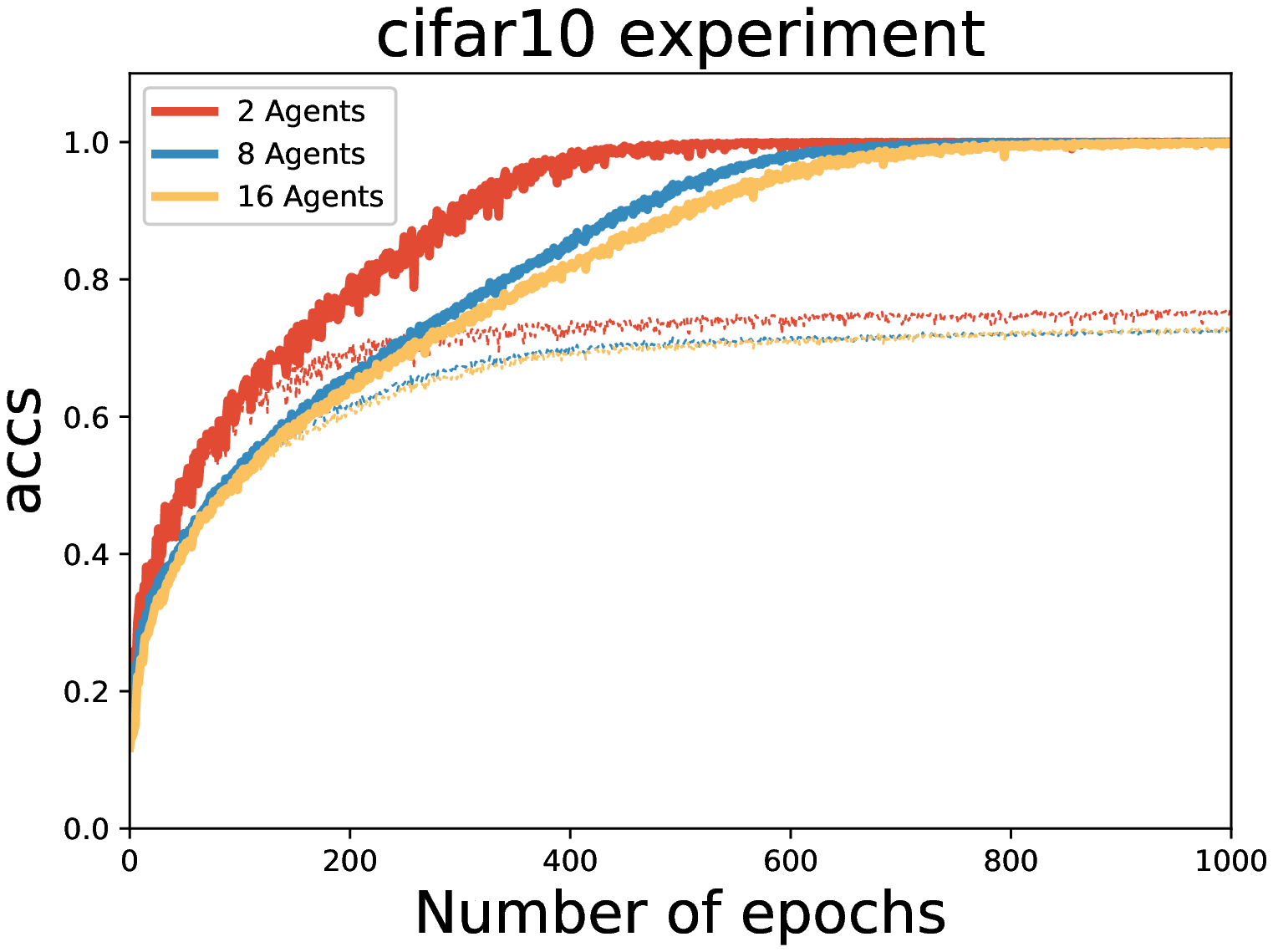}}
\subfigure[]{\includegraphics[width=0.48\textwidth]{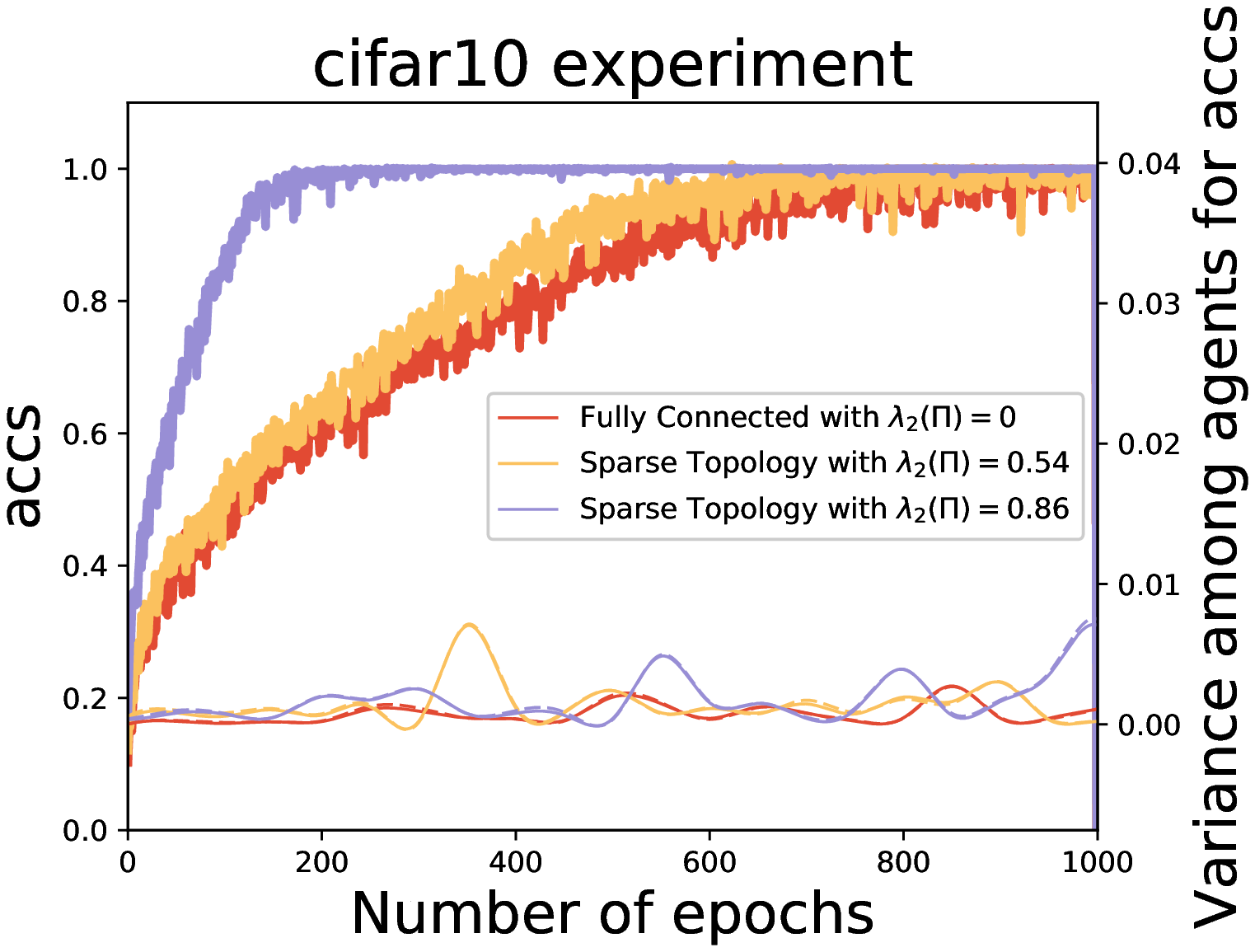}}\vspace{-10pt}
\caption{\textit{Average training (solid lines) and validation (dash lines) accuracy along with accuracy variance over agents for CDMSGD algorithm with (a) varying network size and (b) varying network topology}}\label{Figure2}\vspace{-20pt}
\end{figure}

\vspace{-8pt}
In this section, we investigate the effects of network size and topology on the performance of the proposed algorithms. Figure~\ref{Figure2}(a) shows the change in training performance as the number of agents grow from 2 to 8 and to 16. Although with increase in number of agents, the convergence rate slows down, all networks are able to achieve similar accuracy levels.
%To understand the level of consensus, we also plot the variance of accuracy values over agents (a smooth version using moving average filter) for all the cases and as expected higher level of consensus is achieved quicker for smaller networks. However, all networks eventually achieves high levels of consensus.
Finally, we investigate the impact of network sparsity (as quantified by the second largest eigenvalue) on the learning performance. The primary observation is convergence of average accuracy value happens faster for sparser networks (higher second largest eigenvalue). This is similar to the trend observed for FedAvg algorithm while reducing the Client fraction ($C$) which makes the (stochastic) agent interaction matrix sparser. However, from the plot of the variance of accuracy values over agents (a smooth version using moving average filter), it can be observed that the level of consensus is more stable for denser networks compared to that for sparser networks. This is also expected as discussed in Proposition~\ref{prop1}. Note, with the availability of a central parameter server (as in federated averaging), sparser topology may be useful for a faster convergence, however, consensus (hence, topology density) is critical for a collaborative learning paradigm with decentralized computation.

\vspace{-10pt}
\section{Conclusion and Future Work}\label{conclusion}\vspace{-10pt}
This paper addresses the collaborative deep learning (and many other machine learning) problem in a completely distributed manner (i.e., with data parallelism and decentralized computation) over networks with fixed topology. We establish a consensus based distributed SGD framework and proposed associated learning algorithms that can prove to be extremely useful in practice. Using a Lyapunov function construction approach, we show that the proposed CDSGD algorithm can achieve linear convergence rate with sufficiently small fixed step size and sublinear convergence rate with diminishing step size (see supplementary section~\ref{supp} for details) for strongly convex and Lipschitz differentiable objective functions. Moreover, decaying gradients can be observed for the nonconvex objective functions using CDSGD. Relevant experimental results using benchmark datasets show that CDSGD can achieve centralized SGD level accuracy with sufficient training epochs while maintaining a significantly low generalization error. The momentum variant of the proposed algorithm, CDMSGD can outperform recently proposed FedAvg algorithm which also uses data parallelism but does not perform a decentralized computation, i.e., uses a central parameter server. The effects of network size and topology are also explored experimentally which conforms to the analytical understandings.
%as for the proposed method, no parameter server is required such that local workers can communicate with each other. While FedAvg needs more time for receiving and storing local information and then implements updates.
While current and future research is focusing on extensive testing and validation of the proposed framework especially for large networks, a few technical research directions include: (i) collaborative learning with extreme non-IID data; (ii) collaborative learning over directed time-varying graphs; and (iii) understanding the dependencies between learning rate and consensus.
%\begin{enumerate}
%  \item Dependence of convergence rate on communication period;
%  \item Tradeoff between consensus and disagreement in distributed learning problems; % let's not give up this idea
%  \item Learning problems over directed time-varying graphs.
%\end{enumerate}
%\subsubsection*{Acknowledgments}
\pagebreak
%\section*{References}
\bibliographystyle{unsrt}
\bibliography{NIPS_DisLear}

\newpage

\section{Supplementary Materials for ``Collaborative Deep Learning in Fixed Topology Networks"}\label{supp}
\subsection{Additional analytical results and proofs}
We begin with proofs of the lemmas and theorems that are presented in the main body of the paper without proof. The statements of the lemmas and theorems are presented again for completeness.

\noindent\textbf{Lemma~\ref{lemma_1}}:
Let Assumptions~\ref{assump_objective} and~\ref{assump_1} hold. The iterates of CDSGD (Algorithm~\ref{CDSGD}) satisfy the following inequality $\forall k\in \mathbb{N}$:
\begin{equation}
\mathbb{E}[V(\x_{k+1})]-V(\x_k) \leq -\alpha_k\nabla V(\x_k)^T\mathbb{E}[\nabla \mathcal{J}^i(\x_k)]+\frac{\hat{\gamma}}{2}\alpha_k^2\mathbb{E}[\|\nabla \mathcal{J}^i(\x_k)\|^2]
\end{equation}
\begin{proof}
By Assumption~\ref{assump_objective}, the iterates generated by CDSGD satisfy
\begin{equation}
\begin{aligned}
V(\x_{k+1})-V(\x_k)&\leq \nabla V(\x_k)^T(\x_{k+1}-\x_k)+\frac{1}{2}\hat{\gamma} \|\x_{k+1}-\x_k\|^2\\
                     &=-\alpha\nabla V(\x_k)^T\nabla \mathcal{J}^i(\x_k)+\frac{1}{2}\hat{\gamma}\alpha^2\|\nabla \mathcal{J}^i(\x_k)\|^2\\
\end{aligned}
\end{equation}
Taking expectations on both sides, we can obtain
\begin{equation}
\mathbb{E}[V(\x_{k+1})-V(\x_k)]\leq \mathbb{E}[-\alpha\nabla V(\x_k)^T\nabla \mathcal{J}^i(\x_k)+\frac{1}{2}\hat{\gamma}\alpha^2\|\nabla \mathcal{J}^i(\x_k)\|^2]
\end{equation}
While $V(\x_k)$ is deterministic, $V(\x_{k+1})$ is stochastic due to the random sampling aspect. Therefore, we have
\begin{equation}
\mathbb{E}[V(\x_{k+1})]-V(\x_k)\leq -\alpha\nabla V(\x_k)^T\mathbb{E}[\nabla \mathcal{J}^i(\x_k)]+\frac{1}{2}\hat{\gamma}\alpha^2\mathbb{E}[\|\nabla \mathcal{J}^i(\x_k)\|^2]
\end{equation}
which completes the proof.
\end{proof}

\noindent\textbf{Lemma~\ref{lemma2}}:
Let Assumptions~\ref{assump_objective},~\ref{assump_1}, and~\ref{assumption_2} hold. The iterates of CDSGD (Algorithm~\ref{CDSGD}) satisfy the following inequality $\forall k\in \mathbb{N}$:
\begin{equation}\label{lem2}
\E[V(\x_{k+1})]-V(\x_k)\leq -(\zeta_1-\frac{\hat{\gamma}}{2}\alpha Q_m)\alpha\|\nabla V(\x_k)\|^2+\frac{\hat{\gamma}}{2}\alpha^2Q
\end{equation}
\begin{proof}
Recalling Lemma~\ref{lemma_1} and using Assumption~\ref{assumption_2} and Remark~\ref{rem1}, we have
\begin{equation}
\begin{aligned}
\E[V(\x_{k+1})]-V(\x_k)&\leq -\zeta_1\alpha\|\nabla V(\x_k)\|^2+\frac{\hat{\gamma}}{2}\alpha^2\E[\|\nabla \mathcal{J}^i(\x_k)\|^2]\\
                       &\leq -\zeta_1\alpha\|\nabla V(\x_k)\|^2+\frac{\hat{\gamma}}{2}\alpha^2(Q+Q_m\|\nabla V(\x_k)\|^2)\\
                       %&=-\zeta_1\alpha\|\nabla V(\x_k)\|^2+\frac{\hat{\gamma}}{2}\alpha^2Q+\frac{\hat{\gamma}}{2}\alpha^2\\
                       %&Q_m\|\nabla V(\x_k)\|^2\\
                       &=-(\zeta_1-\frac{\hat{\gamma}}{2}\alpha Q_m)\alpha\|\nabla V(\x_k)\|^2+\frac{\hat{\gamma}}{2}\alpha^2Q
\end{aligned}
\end{equation}
which completes the proof.
\end{proof}

In order to prove Propositon~\ref{prop1}, several auxiliary technical lemmas are presented first.
\begin{lem}\label{lemma_lowerboundofV}
$V$ has a lower bound denoted by $V_{\text{inf}}$ over an open set which contains the iterates $\{\x_k\}$ generated by CDSGD (Algorithm~\ref{CDSGD}).
\end{lem}

Lemma~\ref{lemma_lowerboundofV} can be obtained as each $f_j$ is proper and coercive. Such a lemma is able to help characterize the nonconvex case in which the global optimum may not be achieved.

%\begin{lem}\label{lemma5}
%Let Assumptions~\ref{assump_objective} and~\ref{assump_1} hold. If the step size satisfies that $0\textless \alpha\leq\frac{\zeta_1-(1-\lambda_N(\Pi))Q_m}{\gamma_mQ_m}$, then the iterate $\{\|x_k\|\}$ is bounded above.
%\end{lem}
%\begin{proof}
%Recalling Lemma~\ref{lemma2} can lead to the following:
%\[
%\E[V(\x_{k+1})]-V(\x_k)\leq -(\zeta_1-\frac{\hat{\gamma}}{2}\alpha Q_m)\alpha\|\nabla V(\x_k)\|^2+\frac{\hat{\gamma}}{2}\alpha^2Q
%\]
%With the step size satisfying the above condition, it can be concluded that
%\begin{equation}
%\E[V(\x_{k+1})]-(V(\x_k)+\frac{\hat{\gamma}\alpha^2Q}{2})\leq-\frac{\zeta_1\alpha}{2}\|\nabla V(\x_k)\|^2\leq 0
%\end{equation}
%which implies that the function value sequence $\{V(\x_k)\}$ is bounded above by some positive constants for all $k\in \N$. As $\frac{N}{n}\mathbf{1}^T\F(\x_k)\leq V(\x_k)$ and $V(\x_k)\textless \infty, \forall k\in\N$, then $\frac{N}{n}\mathbf{1}^T\F(\x_k)\textless \infty$, which suggests that the sequence $\{\x_k\}$ is bounded due to the coercivity of $\mathbf{1}^T\F(\x_k)$.
%\end{proof}
\pagebreak
\begin{lem}\label{lemma6}
Let Assumption~\ref{assump_objective} holds. There exists some constant $0\textless L\textless \infty$ such that $\E[\|\mathbf{g}(\x_k)\|]\leq L$.
\end{lem}
The proof of Lemma~\ref{lemma6} directly follows from the Assumption~\ref{assump_objective} c) and $L = \text{max}_jL_j$.
%The proof of Lemma~\ref{lemma6} follows from the Definition~\ref{smooth}, Lemmas~\ref{lemma2} and~\ref{lemma5}. The details of proof are similar as the Lemma 6 in~\cite{zeng2016nonconvex}.

\noindent\textbf{Theorem~\ref{theorem1}}(Convergence of CDSGD with fixed step size, strongly convex case):
Let Assumptions~\ref{assump_objective},~\ref{assump_1} and~\ref{assumption_2} hold. The iterates of CDSGD (Algorithm~\ref{CDSGD}) satisfy the following inequality $\forall k\in \mathbb{N}$, when the step size satisfies
\[
0\textless \alpha\leq\frac{\zeta_1-(1-\lambda_N(\Pi))Q_m}{\gamma_mQ_m}
\]
\begin{equation}
\begin{aligned}
\E[V(\x_k)-V^*]&\leq(1-\alpha \hat{H}\zeta_1)^{k-1}(V(\x_1)-V^*)+\frac{\alpha^2\hat{\gamma}Q}{2}\sum_{l=0}^{k-1}(1-\alpha\hat{H}\zeta_1)^l\\
               &=(1-(\alpha H_m+1-\lambda_2(\Pi))\zeta_1)^{k-1}(V(\x_1)-V^*)\\
               &+\frac{(\alpha^2\gamma_m+\alpha(1-\lambda_N(\Pi)))Q}{2}\sum_{l=0}^{k-1}(1-(\alpha H_m+1-\lambda_2(\Pi))\zeta_1)^{l}
\end{aligned}
\end{equation}
\begin{proof}
Recalling Lemma~\ref{lemma2} and using Eq.~\ref{strong_convexity_formula} yield that
\begin{equation}
\begin{aligned}
\E[V(\x_{k+1})]-V(\x_k)&\leq -(\zeta_1-\frac{\hat{\gamma}}{2}\alpha Q_m)\alpha\|\nabla V(\x_k)\|^2+\frac{\hat{\gamma}}{2}\alpha^2Q\\
                       &\leq -\frac{1}{2}\alpha \zeta_1\|\nabla(\x_k)\|^2+\frac{\alpha^2\hat{\gamma}Q}{2}\\
                       &\leq -\alpha\zeta_1\hat{H}(V(\x_k)-V^*)+\frac{\alpha^2\hat{\gamma}Q}{2}
\end{aligned}
\end{equation}
The second inequality follows from the relation: $\alpha\leq \frac{\zeta_1}{\hat{\gamma}Q_m}$, which is implied by: $\alpha\leq\frac{\zeta_1-(1-\lambda_N(\Pi))Q_m}{\gamma_mQ_m}$. The third inequality follows from the strong convexity. The expectation taken in the above inequalities is only related to $\x_{k+1}$. Hence, recursively taking the expectation and subtracting $V^*$ from both sides requires the following inequality to hold
\begin{equation}\label{thm1}
\E[V(\x_{k+1})-V^*]\leq (1-\alpha\hat{H}\zeta_1)\E[V(\x_k)-V^*]+\frac{\alpha^2\hat{\gamma}Q}{2}
\end{equation}
As $0\textless \alpha\hat{H}\zeta_1\leq\frac{\hat{H}\zeta_1^2}{\hat{\gamma}Q_m}\leq \frac{\hat{H}\zeta_1^2}{\hat{\gamma}\zeta_1^2}=\frac{\hat{H}}{\hat{\gamma}}\leq 1$, the conclusion follows by applying Eq.~\ref{thm1} recursively through iteration $k\in \N$.
\end{proof}

\noindent\textbf{Theorem~\ref{thm2}}(Convergence of CDSGD with fixed step size, nonconvex case):
Let Assumptions~\ref{assump_objective},~\ref{assump_1}, and~\ref{assumption_2} hold. The iterates of CDSGD (Algorithm~\ref{CDSGD}) satisfy the following inequality $\forall m\in \mathbb{N}$, when the step size satisfies
\[
0\textless \alpha\leq\frac{\zeta_1-(1-\lambda_N(\Pi))Q_m}{\gamma_mQ_m}
\]
\begin{equation}
\begin{aligned}
\E[\sum_{k=1}^{m}\|\nabla V(\x_k)\|^2]&\leq \frac{\hat{\gamma}m\alpha Q}{\zeta_1}+\frac{2(V(\x_1)-V_{\text{inf}})}{\zeta_1\alpha}\\
                                      &=\frac{(\gamma_m\alpha+1-\lambda_N(\Pi))mQ}{\zeta_1}+\frac{2(V(\x_1)-V_{\text{inf}})}{\zeta_1\alpha}
\end{aligned}
\end{equation}
%and thus
%\begin{equation}
%\E[\frac{1}{m}\sum_{k=1}^{m}\|\nabla V(\x_k)\|^2]\leq \frac{(\gamma_m\alpha+1-\lambda_N(\Pi))Q}{\zeta_1}+\frac{2(V(\x_1)-V_{\text{inf}})}{m\zeta_1\alpha}
%\end{equation}
\begin{proof}
Recalling Lemma~\ref{lemma2}, and also taking the expectation lead to the following relation,
\begin{equation}\label{nonconvex_descent}
\E[V(\x_{k+1})]-\E[V(\x_k)]\leq -(\zeta_1-\frac{\hat{\gamma}\alpha Q_m}{2})\alpha\E[\|\nabla V(\x_k)\|^2]+\frac{\hat{\gamma}\alpha^2Q}{2}
\end{equation}
As the step size satisfies that $\alpha\leq\frac{\zeta_1}{\hat{\gamma}Q_m}$, it results in
\begin{equation}
\E[V(\x_{k+1})]-\E[V(\x_k)]\leq  -\frac{\zeta_1\alpha}{2}\E[\|\nabla V(\x_k)\|^2]+\frac{\alpha^2\hat{\gamma}Q}{2}
\end{equation}
Applying the above inequality from 1 to $m$ and summing them up can give the following relation
\begin{equation}
V_{\text{inf}} - V(\x_1)\leq \E[V(\x_{k+1})]-V(\x_1)\leq -\frac{\zeta_1\alpha}{2}\sum_{k=1}^{m}\E[\|\nabla V(\x_k)\|^2]+\frac{m\alpha^2\hat{\gamma}Q}{2}
\end{equation}
The last inequality follows from the Lemma~\ref{lemma_lowerboundofV}. Rearrangement of the above inequality and substituting $\hat{\gamma} = \gamma_m+\alpha^{-1}(1-\lambda_N(\Pi))$ into it yield the desired result.
%It is immediately that when both sides are divided by $m$ the second part of results holds.
\end{proof}

\subsection{Proof with Diminishing Step Size}\label{analysis_dim}
From results presented in section~\ref{main_results}, it can be concluded that when the step size is fixed, the function value can only converge near the optimal value. However, in many deep learning models, noisy gradient is quite common due to the random data sampling. Hence, such a situation requires the step size to be adaptive and then with noise, the function value sequence is able to converge to the optimal value. Let $\{\alpha_k\}$ be defined as a diminishing step size sequence that satisfies the following properties:
\[
\alpha_k\textgreater 0,\; \sum_{k=0}^{\infty}\alpha_k = \infty,\; \sum_{k=0}^{\infty}\alpha_k^2\textless \infty
\]
The implication of the above properties is that $\lim_{k\to \infty} \alpha_k = 0$. The next proposition states that when the step size is diminishing, consensus can be achieved asymptotically, i.e., $\lim_{k\to\infty}\E[\|x^j_k-s_k\|]=0$.
\begin{proposition}{(Consensus with diminishing step size)}\label{prop2}
Let Assumptions~\ref{assump_objective} and~\ref{assump_1} hold. The iterates of CDSGD (Algorithm~\ref{CDSGD}) satisfy the following inequality $\forall k\in \mathbb{N}$, when $\alpha_k$ is diminishing,
\begin{equation}
\lim_{k\to\infty}\E[\|x^j_k - s_k\|]=0
\end{equation}
\end{proposition}
The proof is adapted from the Lemma 1 in~\cite{yuan2013convergence}, Lemmas 5 and 6 in~\cite{nedic2015distributed}.

%It can be observed that when the step size is diminishing, the Lyapunov function may approach infinity along with the iterations. Therefore, the Lyapunov function may not be used for addressing the convergence analysis. Moreover, $\|V(\x_k)\|$ approaches infinity as well such that the Assumption~\ref{assumption_2} cannot hold. However, we show that even with diminishing step size, the Lyapunov function and stochastic Lyapunov gradient are finite.

Recalling the algorithm CDSGD
\[
\x_{k+1} = \w_k-\alpha_k\mathbf{g}(\x_k) = \x_k-\alpha_k(\mathbf{g}(\x_k)+\frac{1}{\alpha_k}(\x_k-\w_k))
\]
We define $\nabla \hat{\mathcal{J}}^i(\x_k) =\nabla \mathcal{J}^i(\x_k, \alpha_k)= \mathbf{g}(\x_k)+\frac{1}{\alpha_k}(\x_k-\w_k)$, and the following Lyapunov function
\begin{equation}
\hat{V}(\x)=V(\x, \alpha_k):=\frac{N}{n}\mathbf{1}^T\F(\x)+\frac{1}{2\alpha_k}\|\x\|^2_{I-\Pi}
\end{equation}
The general Lyapunov function is a function of the diminishing step size $\alpha_k$. However, the step size is independent of the variable $\x$ such that it only affects the magnitude of $\|\x\|^2_{I-\Pi}$ along with iterations. Note, from Proposition~\ref{prop2}, we have that each agent eventually reaches the consensus with diminishing step size. Hence, the term $\frac{1}{2\alpha_k}\|\x\|^2_{I-\Pi}$ should not increase with increase in $k$ as the step size $\alpha_k \rightarrow 0$ for $k \rightarrow \infty$.
%We show that when $\alpha_k$ approaches 0 the Lyaponov function turns out to be that $\hat{V}(\x_{\infty}) \to \frac{N}{n}\mathbf{1}^T\F(\x_{\infty})$. Therefore, at the stationary point, no matter it is unique or not, $\hat{V}(\x^*)\to \frac{N}{n}\mathbf{1}^T\F(\x^*)$.
To show that CDSGD with diminishing step size enables convergence to the optimal value, the necessary lemmas and assumptions are directly used from the previous part of the paper with modified constants.

We next show that the Lyapunov function and stochastic Lyapunov gradient with the diminishing step size are bounded. More formally, we aim to show that $\|\nabla\hat{\mathcal{J}}^i(\x_k)\|$ is bounded above for all $k\in \N$. We have, $\|\nabla\hat{\mathcal{J}}^i(\x_k)\| \leq \|\mathbf{g}(\x_k)\|+\frac{1}{\alpha_k}\|(I-\Pi)\x_k\|$ and $\mathbf{g}(\x_k)$ is bounded. Therefore, we have to show that $\frac{1}{\alpha_k}\|(I-\Pi)\x_k\|$ is bounded for all $k\in\N$.
\begin{lem}\label{lemma7}
Let Assumptions~\ref{assump_objective} and~\ref{assump_1} hold. The iterates of CDSGD (Algorithm~\ref{CDSGD}) satisfy the following inequality $\forall k\in \mathbb{N}$, when the step size is diminishing and satisfies that
\[
0\textless\alpha_0\leq \frac{\hat{\zeta}_1-(1-\lambda_N(\Pi))\hat{Q}_m}{\gamma_m\hat{Q}_m},
\]
\begin{equation}
\frac{1}{\alpha_k}\E[\|(I-\Pi)\x_k\|]\textless \infty
\end{equation}
and
\begin{equation}
\lim_{k\to\infty}\E[\|(I-\Pi)\x_k\|] = 0.
\end{equation}
$\hat{\zeta}_1, \hat{Q}_m$ correspond to $\hat{V}$.
\end{lem}
The proof of Lemma~\ref{lemma7} requires another auxiliary technical lemma as follows.
\begin{lem}\label{lemma8}
Let Assumptions~\ref{assump_objective} and~\ref{assump_1} hold. The iterates of CDSGD (Algorithm~\ref{CDSGD}) satisfy the following inequality $\forall k\in \mathbb{N}$, when the step size is diminishing and satisfies that
\[
0\textless\alpha_0\leq \frac{\hat{\zeta}_1-(1-\lambda_N(\Pi))\hat{Q}_m}{\gamma_m\hat{Q}_m},
\]
\begin{equation}
\sum_{k=2}^{\infty}\alpha_k\E[\|(I-\Pi)\x_k\|]\textless \infty.
\end{equation}
\end{lem}
\begin{proof}
Recalling the CDSGD algorithm,
\begin{equation}
\x_{k+1} = \Pi\x_k-\alpha_k\mathbf{g}(\x_k)
\end{equation}
Applying the above equality from 1 to $k-1$ yields that
\begin{equation}
\x_k = \Pi^{k-1}\x_1-\sum_{l=1}^{k-1}\alpha_l\Pi^{k-1-l}\mathbf{g}(\x_l)
\end{equation}
Setting $\x_1 = 0$ results in that $\x_k = -\sum_{l=1}^{k-1}\alpha_l\Pi^{k-1-l}\mathbf{g}(\x_l)$. With this setup, we have
\begin{equation}
\begin{aligned}
\sum_{k=2}^{\infty}\alpha_k\|(I-\Pi)\x_k\|&\leq \sum_{k=2}^{\infty}\alpha_k\|I-\Pi\|\|\x_k\|\\
                                          &\leq \sum_{k=2}^{\infty}\alpha_k\|\sum_{l=1}^{k-1}\alpha_l\Pi^{k-1-l}\mathbf{g}(\x_l)\|\\
                                          &\leq \sum_{k=2}^{\infty}\alpha_k\sum_{l=1}^{k-1}\alpha_l\|\Pi^{k-1-l}\mathbf{g}(\x_l)\|\\
                                          &\leq \sum_{k=2}^{\infty}\alpha_k\sum_{l=1}^{k-1}\alpha_l\|\Pi^{k-1-l}\|\|\mathbf{g}(\x_l)\|\\
                                          &\leq \sum_{k=2}^{\infty}\alpha_k\sum_{l=1}^{k-1}\alpha_l\|\Pi\|^{k-1-l}\|\mathbf{g}(\x_l)\|\\
\end{aligned}
\end{equation}
With the step size being nonincreasing, taking expectation on both side leads to
\begin{equation}
\E[\sum_{k=2}^{\infty}\alpha_k\|(I-\Pi)\x_k\|]\leq \sum_{k=2}^{\infty}\sum_{l=1}^{k-1}\alpha^2_l\|\Pi\|^{k-1-l}\E[\|\mathbf{g}(\x_l)\|]
\end{equation}
As we discussed earlier, we consider that there exists a constant $L$ that bounds $\E[\|\mathbf{g}(\x_k)\|]$ from above for $k\in\N$. Thus, the following relation can be obtained
\begin{equation}
\begin{aligned}
\E[\sum_{k=2}^{\infty}\alpha_k\|(I-\Pi)\x_k\|]&\leq\sum_{k=2}^{\infty}\sum_{l=1}^{k-1}\alpha^2_l\lambda_2(\Pi)^{k-1-l}\E[\|\mathbf{g}(\x_l)\|]\\
                                              &\leq L\sum_{k=2}^{\infty}\sum_{l=1}^{k-1}\alpha^2_l\lambda_2(\Pi)^{k-1-l}
\end{aligned}
\end{equation}
As $\sum_{k=1}^{\infty}\alpha_k^2\textless \infty$ and $\lambda_2(\Pi)\textless 1$ then by Lemma 5 in~\cite{nedic2015distributed}, the desired result follows.
\end{proof}
\begin{proof}[Proof of Lemma~\ref{lemma7}]
We first define that $h_k = \frac{\|(I-\Pi)\x_k\|}{\alpha_k}$. Hence, the result of Lemma~\ref{lemma8} can be rewritten as $\sum_{k=2}^{\infty}\alpha_k^2\E[h_k]\textless \infty$. By defining $h_m = \sup\{\E[h_k]\}$, we have $h_m\sum_{k=2}^{\infty}\alpha_k^2\textless \infty$, which implies that $h_m\textless \infty$ as $\sum_{k=1}^{\infty}\alpha^2_k\textless \infty$. Hence, it is immediately seen that $\E[h_k]\textless \infty$. As $k\to\infty, \alpha_k\to 0$ and $\frac{1}{\alpha_k}\E[\|(I-\Pi)\x_k\|]\textless \infty$, then $\lim_{k\to\infty}\E[\|(I-\Pi)\x_k\|] = 0$, which completes the proof.
\end{proof}
The implication of Lemma~\ref{lemma7} is two folds: One can observe that $\hat{V}(\x_k)$ and $\nabla\hat{\mathcal{J}}^i(\x_k)$ are finite even with diminishing step size such that based on Definition~\ref{smooth} there exists a finite positive constant $\gamma'$ to allow the smoothness of $\hat{V}(\x_k)$ for all $k\in \N$ to hold true; another observation is that Assumption~\ref{assumption_2} still can be used in the main results. It can also be concluded that $\hat{V}$ is strongly convex with some constant $0\leq H'\textless \infty$, where $H'$ corresponds to $\hat{V}$.
\begin{thm}\label{thm3}{(Convergence of CDSGD with diminishing step size, strongly convex case)}
Let Assumptions~\ref{assump_objective},~\ref{assump_1} and~\ref{assumption_2} hold. The iterates of CDSGD (Algorithm~\ref{CDSGD}) satisfy the following inequality $\forall k\in \mathbb{N}$, when the step size is diminishing and satisfies that
\[
0\textless\alpha_0\leq \frac{\hat{\zeta}_1-(1-\lambda_N(\Pi))\hat{Q}_m}{\gamma_m\hat{Q}_m},
\]
\begin{equation}
\begin{aligned}
\E[\hat{V}(\x_k)-\hat{V}^*]&\leq\beta^{k-2}(\hat{V}(\x_1)-\hat{V}^*)+\frac{\gamma'\hat{Q}}{2}\sum_{p=1}^{k-2}\beta^{k-p-2}\alpha^2_p\\
                           &+\frac{\gamma'\hat{Q}\alpha_{k-1}^2}{2}
\end{aligned}
\end{equation}
where $\text{sup}\{1-\alpha_kH'\hat{\zeta}_1\}\leq \beta\textless 1$, and $\gamma', \hat{Q}$ correspond to $\hat{V}$.
\end{thm}
\begin{proof}
As $\alpha_1\leq \frac{\hat{\zeta}_1}{\gamma'\hat{Q}_m}$, then it can be obtained that $\alpha_k\gamma'\hat{Q}_m\leq \alpha_1\gamma'\hat{Q}_m\leq \hat{\zeta}_1$ for all $k\in\N$. Recalling Lemma~\ref{lemma2} and Eq.~\ref{strong_convexity_formula}, subtracting $\hat{V}^*$ from both sides, and taking the expectation yield the following relation
\begin{equation}
\E[\hat{V}(\x_{k+1})- \hat{V}^*]\leq (1-\alpha_kH'\hat{\zeta}_1)\E[\hat{V}(\x_k)-\hat{V}^*]+\frac{\gamma'\hat{Q}\alpha_k^2}{2}
\end{equation}
Applying the above inequality recursively can give the following relation
\begin{equation}
\begin{aligned}
\E[\hat{V}(\x_{k+1})- \hat{V}^*]&\leq (1-\alpha_kH'\hat{\zeta}_1)(1-\alpha_{k-1}H'\hat{\zeta}_1)\E[\hat{V}(\x_{k-1})-\hat{V}^*]\\
                                &+(1-\alpha_{k-1}H'\hat{\zeta}_1)\frac{\gamma'\hat{Q}\alpha_{k-1}^2}{2}+\frac{\gamma'\hat{Q}\alpha_k^2}{2}\\
\end{aligned}
\end{equation}
By induction, the following can be obtained
\begin{equation}
\begin{aligned}
&\E[\hat{V}(\x_{k+1})- \hat{V}^*]\leq\\ &\prod_{q=1}^{k}(1-\alpha_qH'\hat{\zeta}_1)(\hat{V}(\x_1)-\hat{V}^*)+\frac{\gamma'\hat{Q}}{2}\sum_{p=1}^{k-1}\prod_{r=p+1}^{k}(1-\alpha_rH'\hat{\zeta}_1)\alpha_p^2\\
&+\frac{\gamma'\hat{Q}\alpha_{k}^2}{2}
\end{aligned}
\end{equation}
As $H'\leq \gamma'$, it can be derived that $0\textless\alpha_kH'\hat{\zeta}_1\leq 1$ for all $k\in\N$. Therefore, $1-\alpha_kH'\hat{\zeta}_1\in [0,1)$ such that we can define a positive constant $\beta$ satisfies that $\text{sup}\{1-\alpha_kH'\hat{\zeta}_1\}\leq \beta\textless 1$. Hence, combining the last inequalities together, we have
\begin{equation}
\begin{aligned}
\E[\hat{V}(\x_{k+1})- \hat{V}^*]&\leq\beta^{k-1}(\hat{V}(\x_1)-\hat{V}^*)+\frac{\gamma'\hat{Q}}{2}\sum_{p=1}^{k-1}\beta^{k-p-1}\alpha^2_p\\
          &+\frac{\gamma'\hat{Q}\alpha_{k}^2}{2}
\end{aligned}
\end{equation}
which completes the proof by replacing $k+1$ with $k$.
\end{proof}
\begin{rem}
From Theorem~\ref{thm3}, we can conclude that the function value sequence $\{\hat{V}(\x_k)\}$ asymptotically converges to the optimal value. (This holds regardless of whether the ``gradient noise" parameter $\hat{Q}$ is zero or not.) In fact, we can establish the rate of convergence as follows: the first term on the right hand side decreases exponentially if $\beta.< 1$, and the last term decreases as quickly as $\alpha_k^2$. For the middle term, we can use Lemma 3.1 of \cite{ram2012} that establishes bounds on the convolution of two scalar sequences. If we choose $t\textgreater 0$ such that $\alpha_k = \frac{1}{k^\epsilon+t}$, where $\epsilon\in(0.5,1]$, then the necessary growth conditions on $\alpha_k$ are satisfied; substituting this into Theorem~\ref{thm3} yields the stated convergence rate of $\mathcal{O}(\frac{1}{k^\epsilon})$. In practice, $\alpha_k$ can be made adaptive to $\frac{\Theta}{k^\epsilon+t}$ for any constant $\Theta\textgreater 0$.
\end{rem}
Similarly, we also present the convergence results for the nonconvex objective functions.
\begin{thm}\label{thm4}{(Convergence of CDSGD with diminishing step size, nonconvex case)}
Let Assumptions~\ref{assump_objective},~\ref{assump_1} and~\ref{assumption_2} hold. The iterates of CDSGD (Algorithm~\ref{CDSGD}) satisfy the following inequality $\forall m\in \mathbb{N}$, when the step size is diminishing and satisfies that
\[
0\textless\alpha_0\leq \frac{\hat{\zeta}_1-(1-\lambda_N(\Pi))\hat{Q}_m}{\gamma_m\hat{Q}_m},
\]
\begin{equation}\label{diminishing_nonconvex}
\begin{aligned}
\E[\sum_{k=1}^{m}\alpha_k\|\nabla\hat{V}(\x_k)\|^2]\leq \frac{2(\hat{V}(\x_1)-\hat{V}_{inf})}{\hat{\zeta}_1}
                                           +\frac{\gamma'\hat{Q}}{\hat{\zeta}_1}\sum_{k=1}^{m}\alpha_k^2
\end{aligned}
\end{equation}
%and thus
%\begin{equation}
%\lim_{m\to\infty}\E[\frac{1}{S_m}\sum_{k=1}^{m}\alpha_k\|\nabla\hat{V}(\x_k)\|^2]\to 0
%\end{equation}
%where $S_m=\sum_{k=1}^{m}\alpha_k$.
\end{thm}
\begin{proof}
Assume that $\alpha_k\gamma'\hat{Q}_m\leq \hat{\zeta}_1$ for all $k\in\N$. Based on Eq.~\ref{nonconvex_descent} we consider the diminishing step size and Lyapunov function, then the following relation can be obtained
\begin{equation}
\E[\hat{V}(\x_{k+1})]-\E[\hat{V}(\x_k)]\leq-(\hat{\zeta}_1-\frac{\gamma'\alpha_k\hat{Q}_m}{2})\alpha_k\E[\|\nabla\hat{V}(\x_k)\|^2]+\frac{\gamma'\alpha_k\hat{Q}}{2}
\end{equation}
Combining the condition for the step size yields the following inequality
\begin{equation}
\E[\hat{V}(\x_{k+1})]-\E[\hat{V}(\x_k)]\leq-\frac{\hat{\zeta}_1\alpha_k}{2}\E[\|\nabla\hat{V}(\x_k)\|^2]+\frac{\gamma'\alpha_k\hat{Q}}{2}
\end{equation}
Applying the last inequality from 1 to $m$ and summing them up,
\begin{equation}
\begin{aligned}
\hat{V}_{inf}-\E[\hat{V}(\x_1)]&\leq \E[\hat{V}(\x_{k+1})]-\E[\hat{V}(\x_1)]\leq\\
                               &-\frac{\hat{\zeta}_1}{2}\sum_{k=1}^{m}\alpha_k\E[\|\nabla\hat{V}_(\x_k)\|^2]+\frac{\gamma'\hat{Q}}{2}\sum_{k=1}^{m}\alpha_k^2
\end{aligned}
\end{equation}
Dividing by $\hat{\zeta}_1/2$ and rearranging the terms lead to the desired results.
%\begin{equation}
%\sum_{k=1}^{m}\alpha_k\E[\|\nabla\hat{V}(\x_k)\|^2]\leq \frac{2(\E[\hat{V}(\x_1)]-\hat{V}_{inf})}{\hat{\zeta}_1}+\frac{\gamma'\hat{Q}}{\hat{\zeta}_1}\sum_{k=1}^{m}\alpha_k^2
%\end{equation}
\end{proof}
\begin{rem}
Compared to Theorem~\ref{thm2}, Theorem~\ref{thm4} has shown the decaying of gradient $\|\nabla\hat{V}(\x_k)\|$ even with noise when the step size is diminishing in the nonconvex case. This is because when $k\to\infty$, the right hand side of Eq.~\ref{diminishing_nonconvex} remains finite such that $\|\nabla\hat{V}(\x_k)\|^2$ approaches 0.
\end{rem}
\subsection{Additional pseudo-codes of the algorithms}
Momentum methods have been regarded as effective methods to speed up the convergence in numerous optimization problems. While the Nesterov Momentum method has been extended widely to generate variants with provable global convergence properties, the global convergence analysis of Polyak Momentum methods is still quite challenging and an active research topic. Pseudo-codes of CDSGD combined with Polyak momentum and Nesterov momentum methods are presented below.\\

\begin{algorithm}[H]\label{CDSGD-P}
    \caption{CDSGD with Polyak Momentum}
    \SetKwInOut{Input}{Input}
    \SetKwInOut{Output}{Output}

    \Input{$m$, $\alpha$, $N$, $\mu$ (momentum term)}
    \textbf{Initialize}:\text{$x^j_0$, $v^j_0$}\\
%    \Output{Optimizer $x^*$}
    \text{Distribute the training data set to $N$ agents}\\
    \text{For each agent:}\\
    \For{$k=0:m$}
      {
        \text{Randomly shuffle the corresponding data subset}\;
        $w_{k+1}^j = \sum_{l\in Nb(j)}\pi_{jl}x^l_{k}$\\
        $v^j_{k+1} = \mu v^j_k - \alpha_kg_j(x_k^j)$\\
        $x^j_{k+1} = w_{k+1}^j + v^j_{k+1}$
      }
\end{algorithm}

\begin{algorithm}[H]\label{CDSGD-N}
    \caption{CDSGD with Nesterov Momentum}
    \SetKwInOut{Input}{Input}
    \SetKwInOut{Output}{Output}

    \Input{$m$, $\alpha$, $N$, $\mu$}
    \textbf{Initialize}:\text{$x^j_0$, $v^j_0$}\\
%    \Output{Optimizer $x^*$}
    \text{Distribute the training data set to $N$ agents}\\
    \text{For each agent:}\\
    \For{$k=0:m$}
      {
        \text{Randomly shuffle the corresponding data subset}\\
        $w_{k+1}^j = \sum_{l\in Nb(j)}\pi_{jl}x^l_{k}$\\
        $v^j_{k+1} = \mu v^j_k - \alpha_kg_j(x_k^j+\mu v^j_k)$\\
        $x^j_{k+1} = w_{k+1}^j + v^j_{k+1}$
      }
\end{algorithm}

%\begin{algorithm}[H]\label{FedAvg}
%    \caption{FedAvg}
%    \SetKwInOut{Input}{Input}
%    \SetKwInOut{Output}{Output}
%
%    \Input{$m$, $x^j_0$, $\{\alpha_k\}(k=0,1,\dots,m)$, $ g_j(x_0^j)$, $N$, $n$, $n_j(j=1,2,\dots,N)$, $\tilde{x}_0$}
%    \Output{Optimizer $x^*$}
%    \text{Localize $N$ agents for the whole task}\;
%    \text{Distribute the training data set to $N$ agents}\;
%    \text{For each agent:}\\
%    \For{$k=0:m$}
%      {
%        \text{Randomly shuffle the corresponding data subset}\;
%        $x^j_{k+1} = \tilde{x}_k - \alpha_kg_j(x^j_k)$\\
%%            $x = x_k^j - \eta(x^j_k-\tilde{x}_k)$\\
%        $\tilde{x}_{k+1} =\sum_{j=1}^{N}\frac{n_j}{n}x^j_{k+1}$
%      }
%\end{algorithm}

%\begin{algorithm}[H]\label{EAMSGD}
%    \caption{EAMSGD}
%    \SetKwInOut{Input}{Input}
%    \SetKwInOut{Output}{Output}
%
%    \Input{$m$, $x^j_0$, $\{\alpha_k\}(k=0,1,\dots)$, $ g_j(x_0^j + \beta v^j_0)$, $N$, $\tau$, $\beta$, $\eta$, $\tilde{x}_0$, $v_0^j$, $\rho$}
%    \Output{Optimizer $x^*$}
%    \text{Localize $N$ agents for the whole task}\;
%    \text{Distribute the training data set to $N$ agents}\;
%    \text{For each agent:}\\
%    \For{$k=0:m$}
%      {
%        \text{Randomly shuffle the corresponding data subset}\;
%        %$x = x^j_k$\\
%        \If {$k\equiv 0$ \text{mod} $\tau$}
%          {
%            $x = x_k^j - \eta(x^j_k-\tilde{x}_k)$\\
%            $\tilde{x}_{k+1} = (1-\rho)\tilde{x}_k + \rho\frac{1}{N}\sum_{j=1}^{N}x^j_k$
%          }
%        $v^j_{k+1} = \beta v^j_k - \alpha_kg_j(x_k^j + \beta v^j_k)$\\
%        $x^j_{k+1} = x + v^j_{k+1}$
%      }
%\end{algorithm}

\subsection{Additional Experimental Results}
We begin with a discussion on the training loss profiles for the CIFAR-10 results presented in the main body of the paper.
%We perform the experiments by shuffling the data and partitioning it equally among the various agents to ensure that all the agents have IID data.
% Do we need to write about the coding backend and the availability of the code later. If so please uncomment the below part.
% The training performed using Keras and TensorFlow. The code shall be available soon in github.

%uncomment this in case you want to write these details here instead of top
%In Federated Averaging (FASGD) Algorithm, we choose $E$=1 and $C$=1.0 as the parameters in the FASGD algorithm as it is close to a fully connected topology of the agents as considered in the CDSGD and CDMSGD experiments. Also, it can be seen in Section \ref{network_variation_experiments} that the conclusions obtained from this work can be extrapolated to various parameters of the FASGD algorithm.

\begin{figure}
\centering
\subfigure[]{\includegraphics[width=0.48\textwidth]{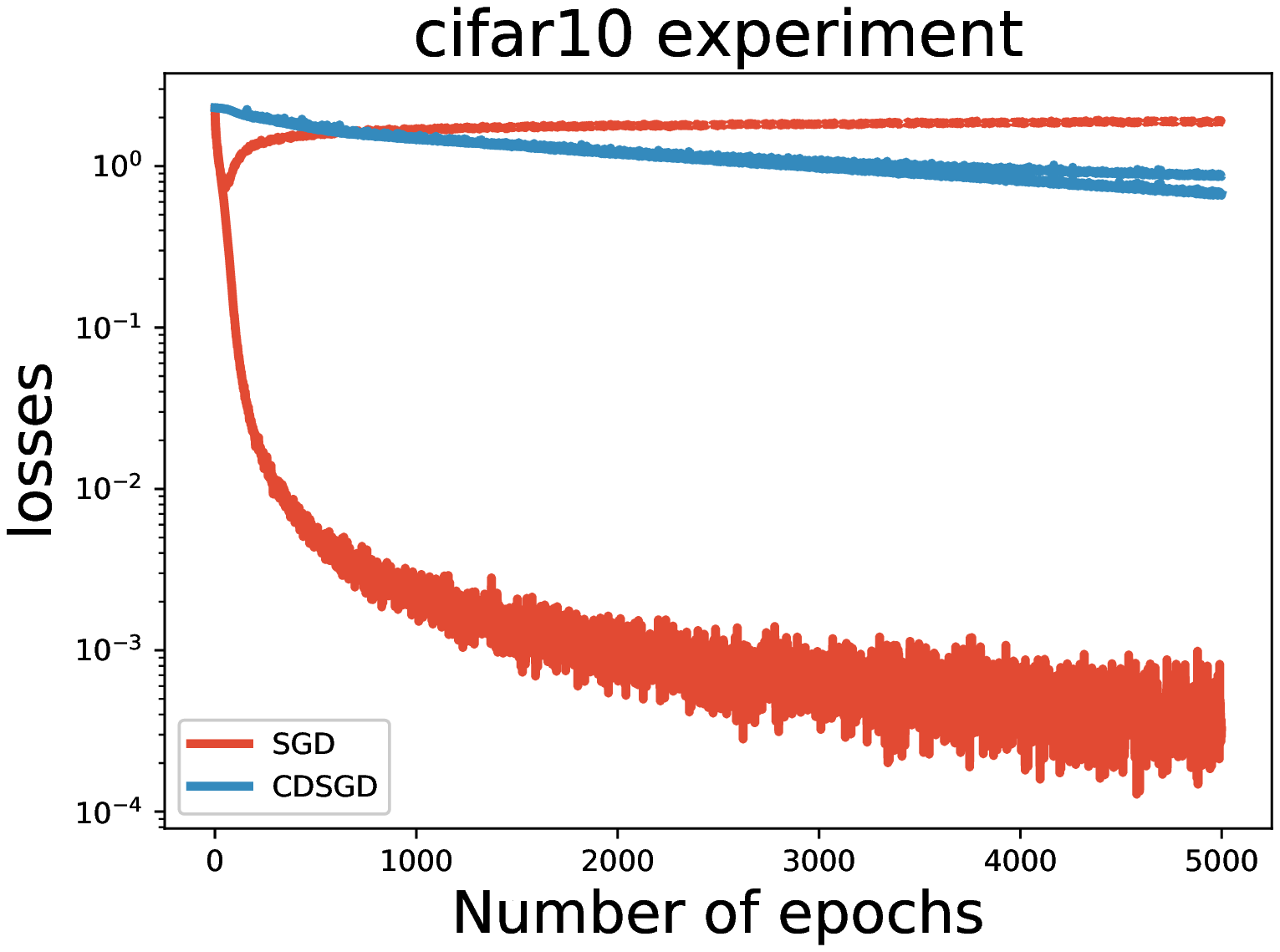}}
\subfigure[]{\includegraphics[width=0.48\textwidth]{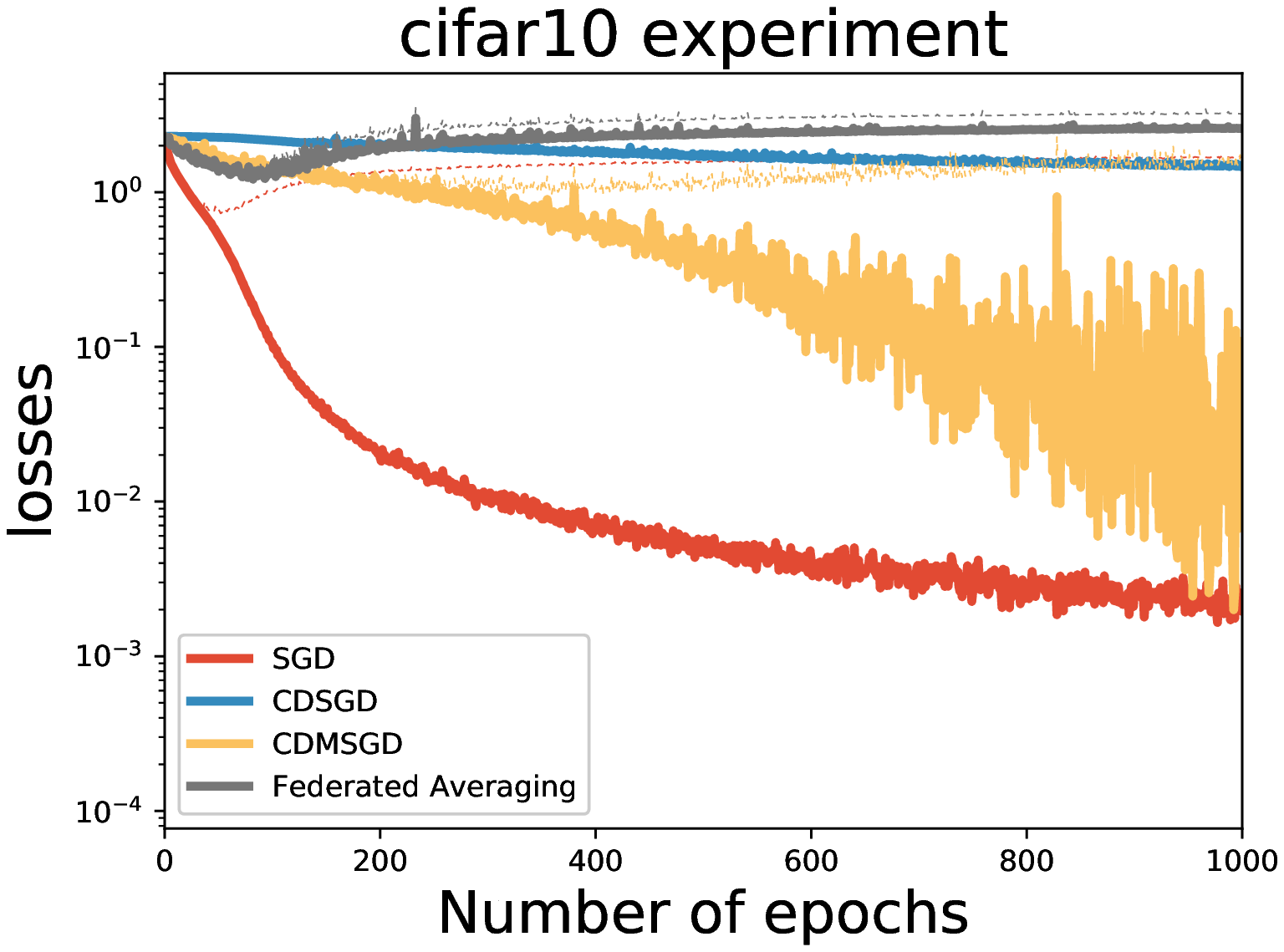}}
\caption{\textit{Average training (solid lines) and validation (dash lines) loss for (a) CDSGD algorithm with SGD algorithm and (b) CDMSGD with Federated averaging method}}\label{Figure3}
\end{figure}

\subsubsection{Comparison of the loss for benchmark methods}
Figure~\ref{Figure3} (a) shows the loss (in log scale) with respect to the number of epochs for SGD and CDSGD algorithms. The solid curve means training and the dash curve indicates validation. From the loss results, it can be observed that SGD has the sublinear convergence rate for training and dominates among the two methods during the training process. While for the validation, SGD performs poorly after around 70 epochs. However, CDSGD shows linear convergence rate (in log scale as discussed in the analysis) for both training and validation. Though, it takes a lot of more time compared to SGD for convergence, it eventually performs better than SGD in the validation data and the gap between the training and validation loss (i.e., the generalization gap~\cite{ZBHR16}) is very less compared to that in SGD.

\subsubsection{Results on CIFAR-100 dataset}
\begin{figure}
\centering
\subfigure[]{\includegraphics[width=0.48\textwidth]{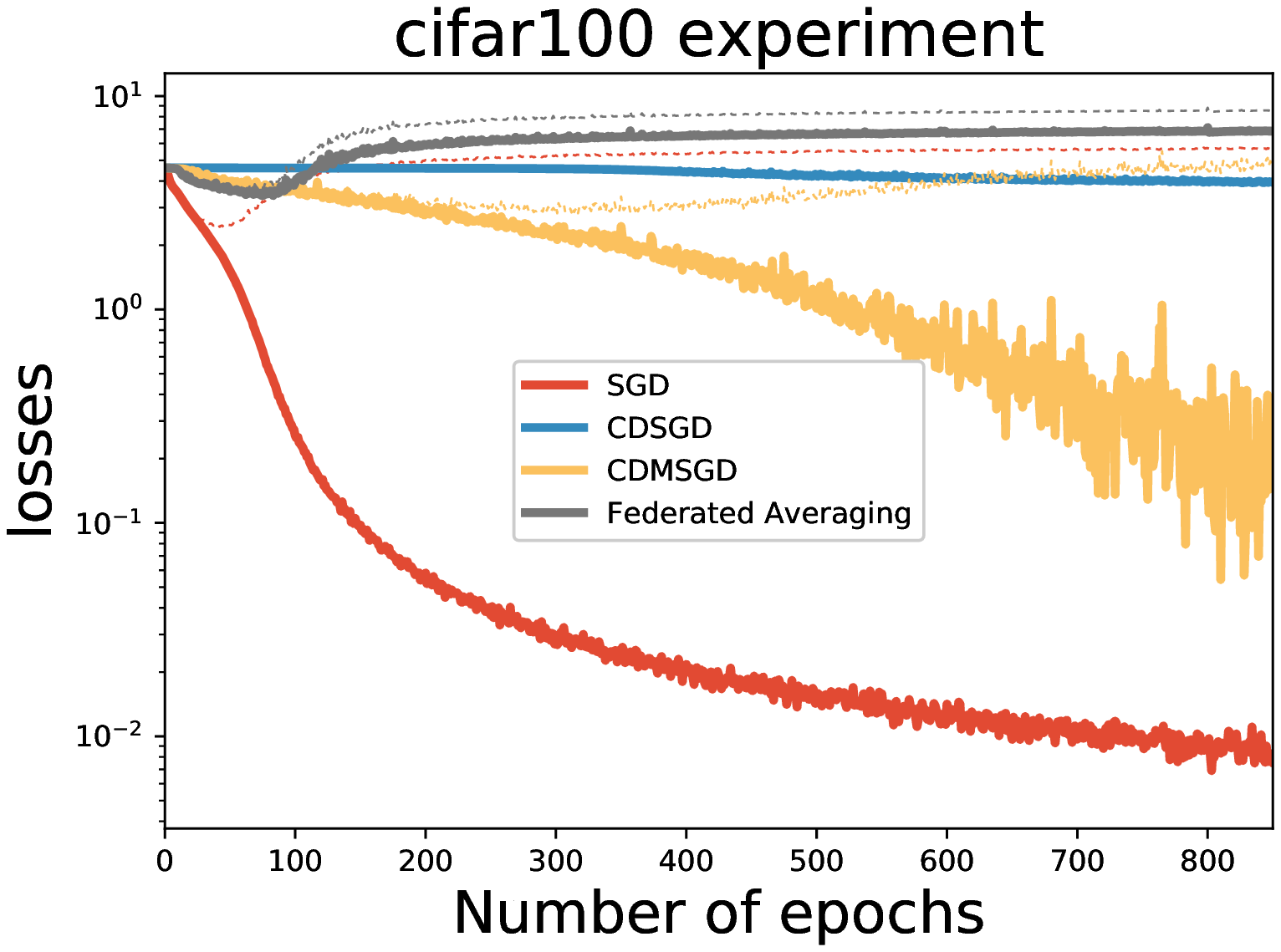}}
\subfigure[]{\includegraphics[width=0.48\textwidth]{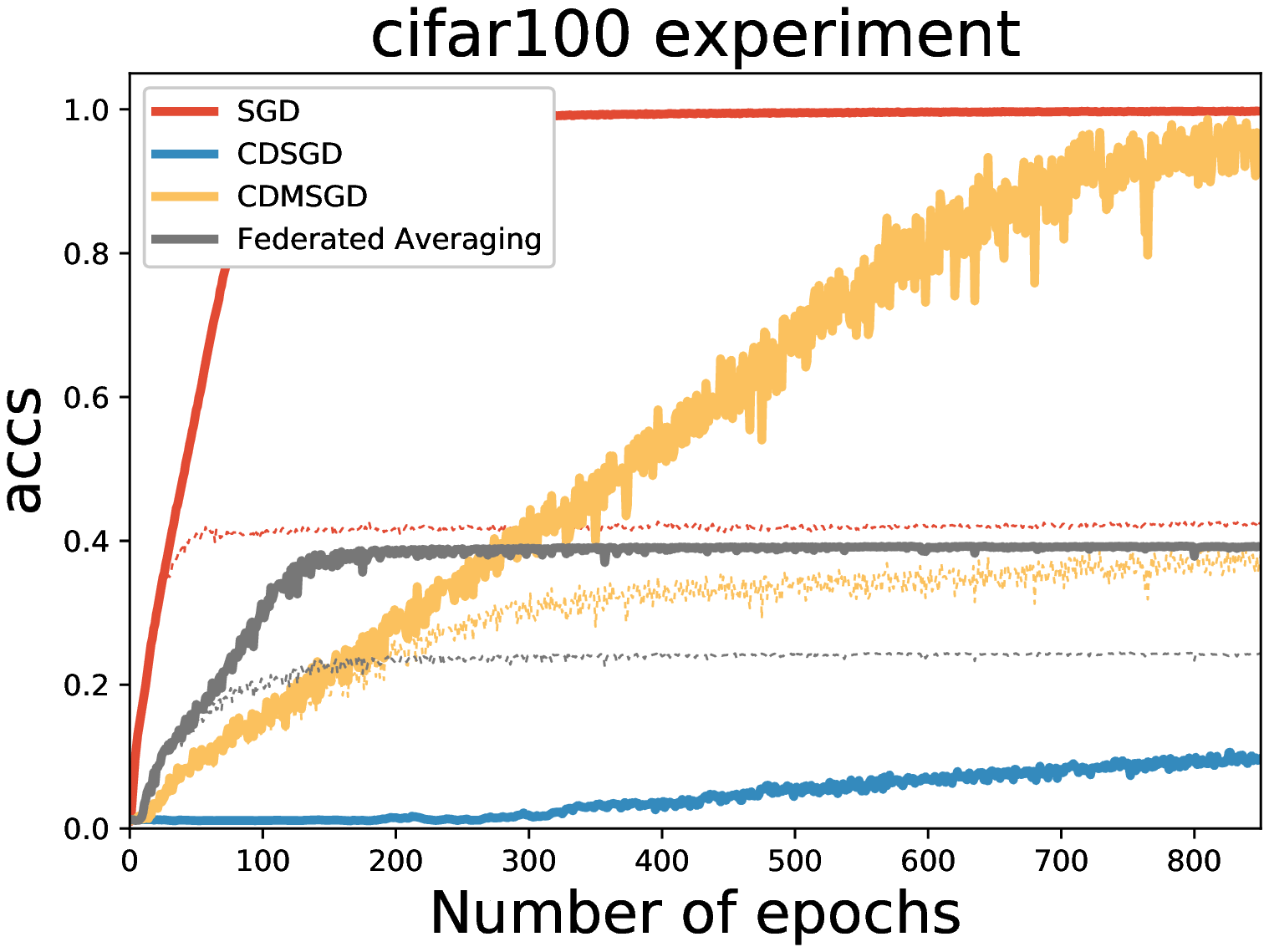}}
\subfigure[]{\includegraphics[width=0.48\textwidth]{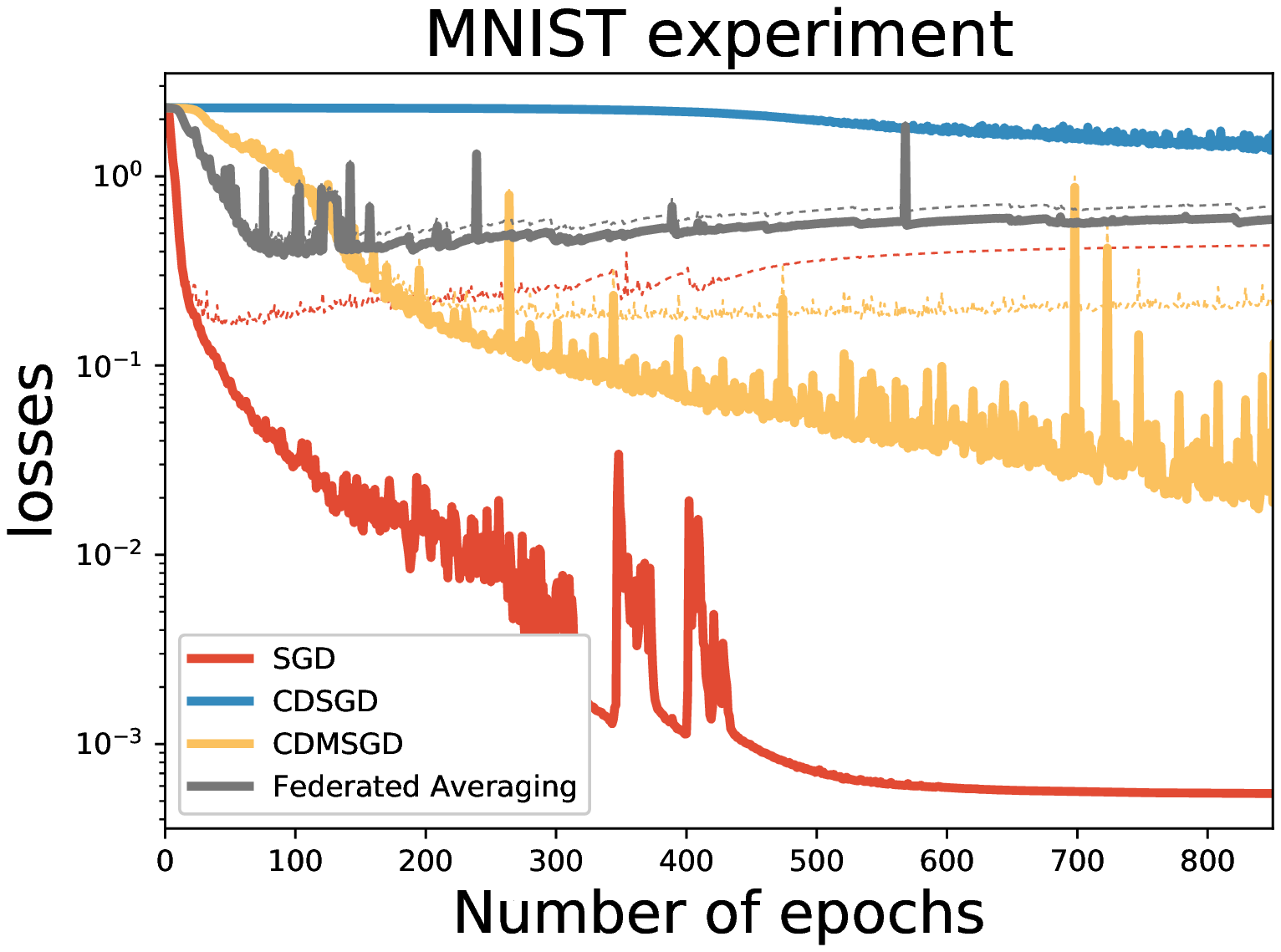}}
\subfigure[]{\includegraphics[width=0.48\textwidth]{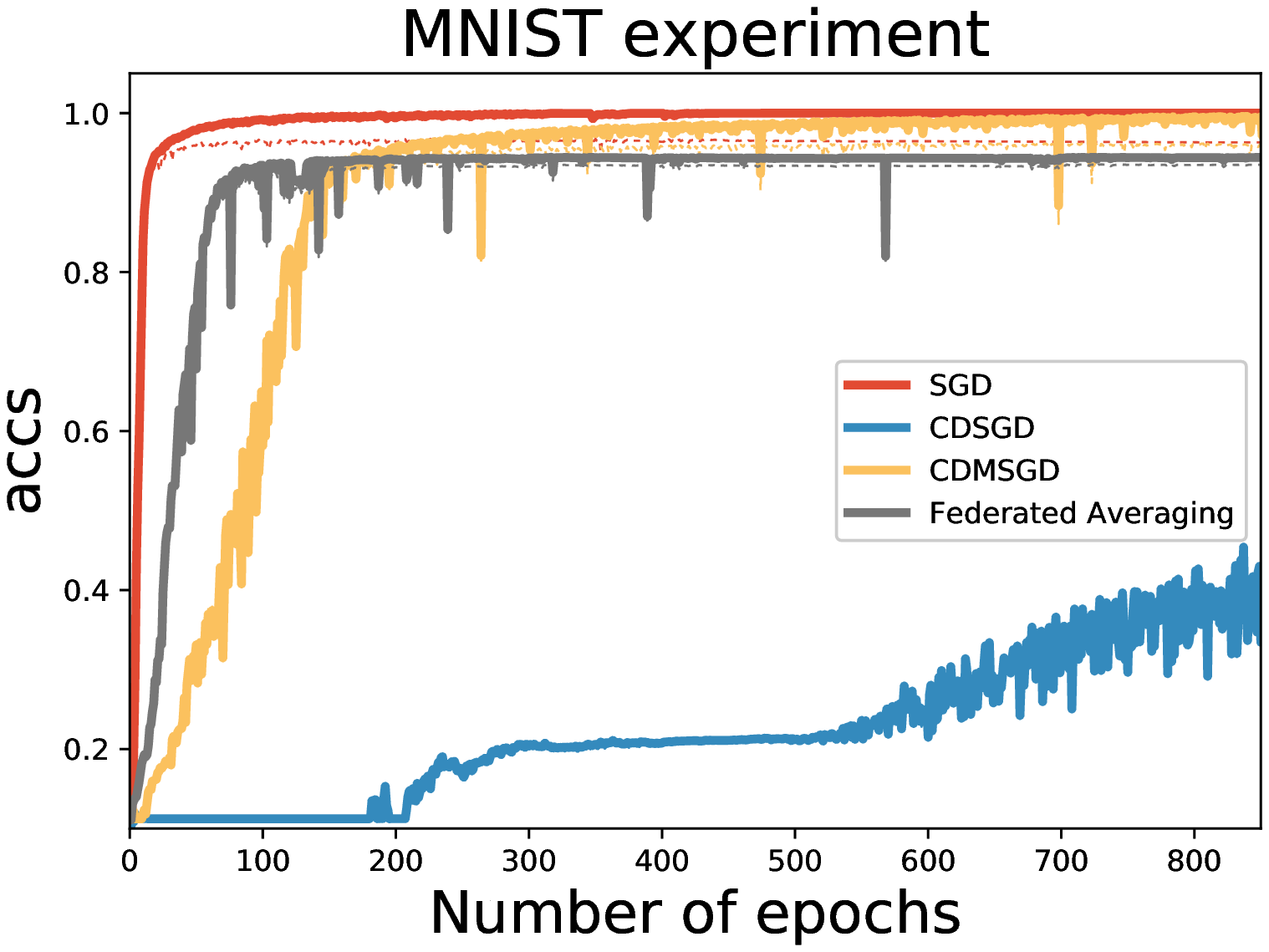}}
\caption{\textit{Average training (solid lines) and validation (dash lines) (a) loss and (b) accuracy for SGD, CDSGD, CDMSGD and Federated averaging method for the CIFAR-100 dataset  (c) loss and (d) accuracy for SGD, CDSGD, CDMSGD and Federated averaging method for the MNIST dataset}}\label{Figure4}
\end{figure}

For the experiments on the CIFAR-100 dataset, we use a CNN similar to that used for the CIFAR-10 dataset. While the results of CIFAR-100 also converges fast for SGD, CDMSGD and Federated Averaging SGD (FedAvg) algorithms (CDMSGD being the slowest) as shown in Figure~\ref{Figure4}, it can be seen that eventually, the loss converges better than the FedAvg algorithm. Similar to the observation made for the CIFAR-10 dataset, we observe that CDMSGD achieves significantly higher validation accuracy compared to FedAvg while approaching similar accuracy level as that of (centralized) SGD. It can also be seen that as expected CDSGD's convergence is very slow compared to the others.

\subsubsection{Results on MNIST dataset}
For the experiments on the MNIST dataset, the model used for training is a Deep Neural Network with 20 Fully Connected layers consisting of 50 ReLU units each and the output layer with 10 units having softmax activation. The model was trained using the catagorical cross-entropy loss. Figure ~\ref{Figure4}(c \& d) shows the loss and accuracy obtained over the number of epochs. In this case, while the accuracy levels are significantly higher as expected for the MNIST dataset, the trends remain consistent with the results obtained for the other benchmark datasets of CIFAR-10 and CIFAR-100. Note, the generalization gap between the training and validation data for all the methods are very less (least for CDMSGD).

\subsubsection{Effect of the decaying step size}
Based on the analysis presented in section~\ref{analysis_dim}, it is evident that decaying step size has a significant effect on the accuracy as well as convergence. A performance comparison of SGD, Momentum SGD (MSGD) and CDMSGD with a decaying stepsize is performed using the MNIST dataset. It can be seen that the performance of the CDMSGD with decaying step size becomes slightly better than SGD with decaying step size while (centralized) MSGD has the best performance. Although CDMSGD sometimes suffers from large fluctuations, it demonstrates the least generalization gap among all the algorithms.

\begin{figure}
\centering
\subfigure[]{\includegraphics[width=0.48\textwidth]{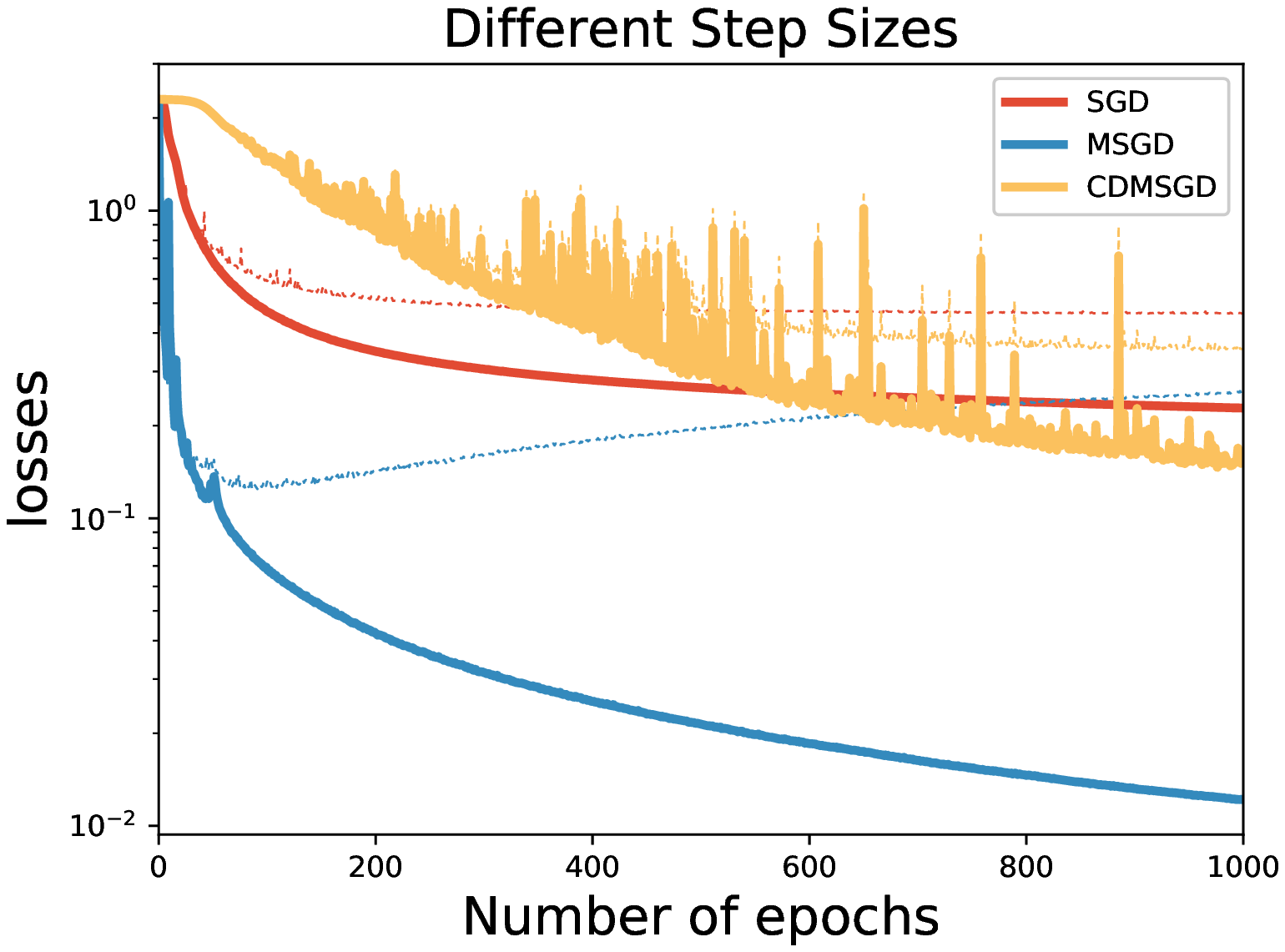}}
\subfigure[]{\includegraphics[width=0.48\textwidth]{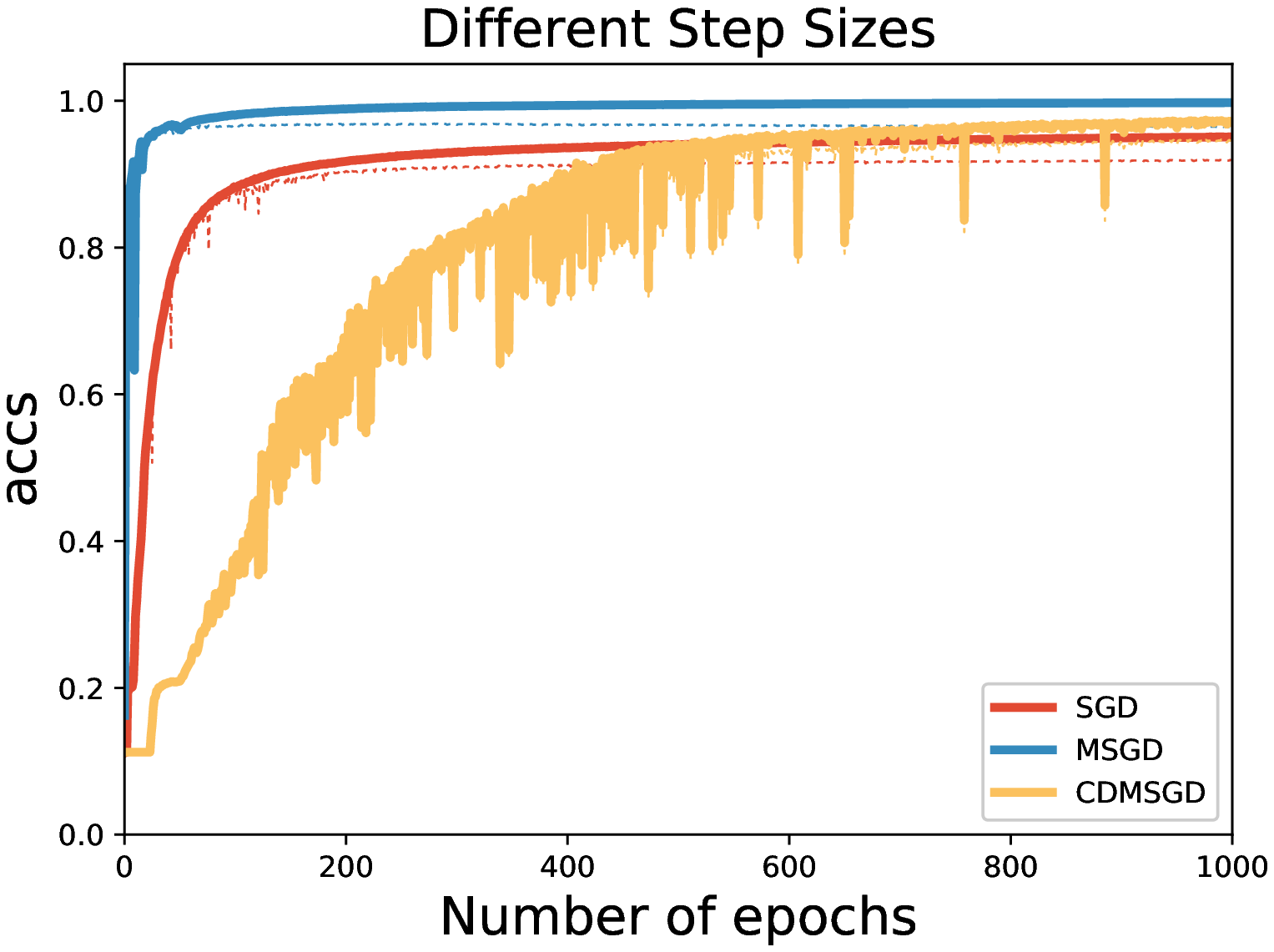}}
\subfigure[]{\includegraphics[width=0.48\textwidth]{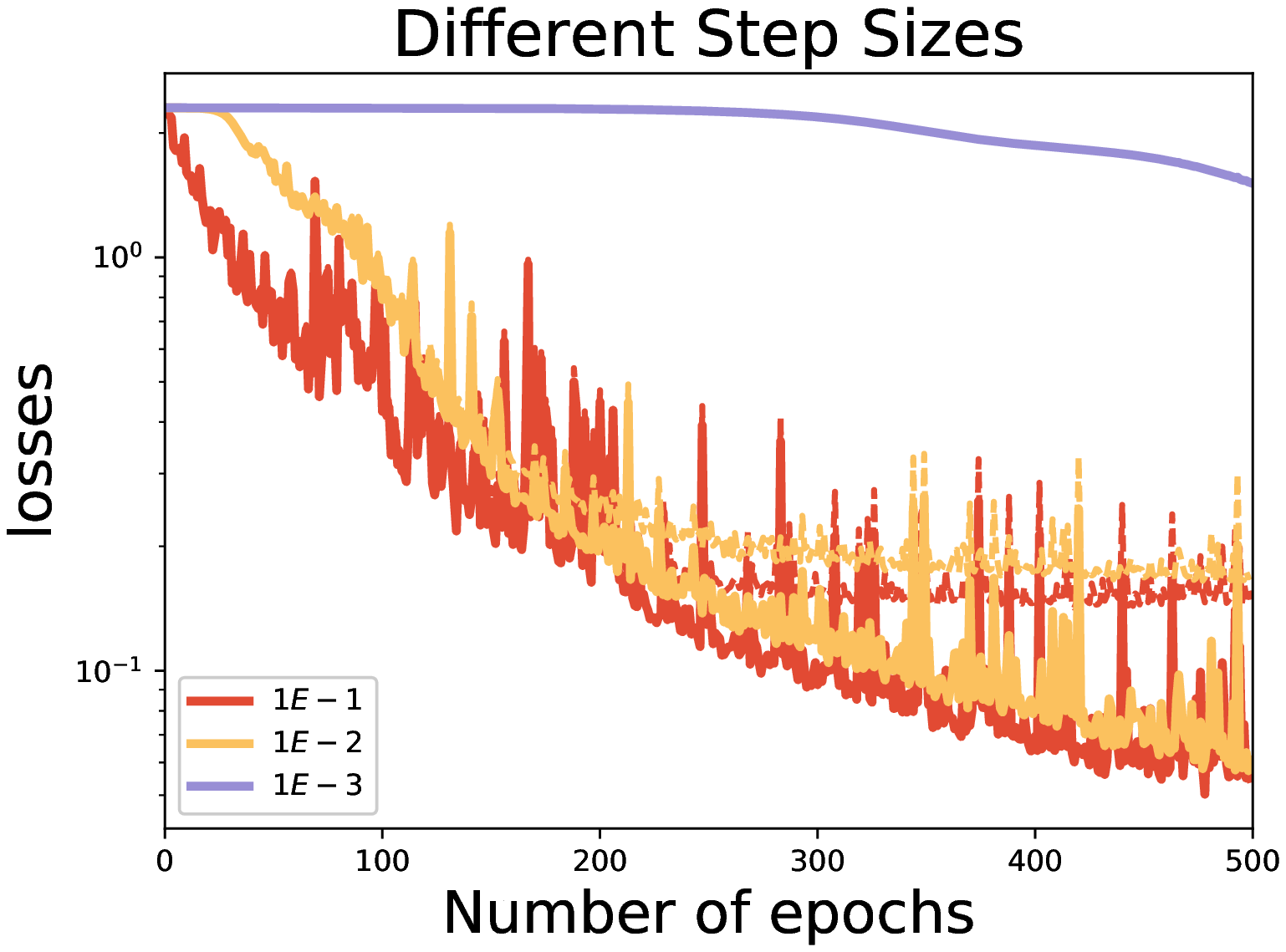}}
\subfigure[]{\includegraphics[width=0.48\textwidth]{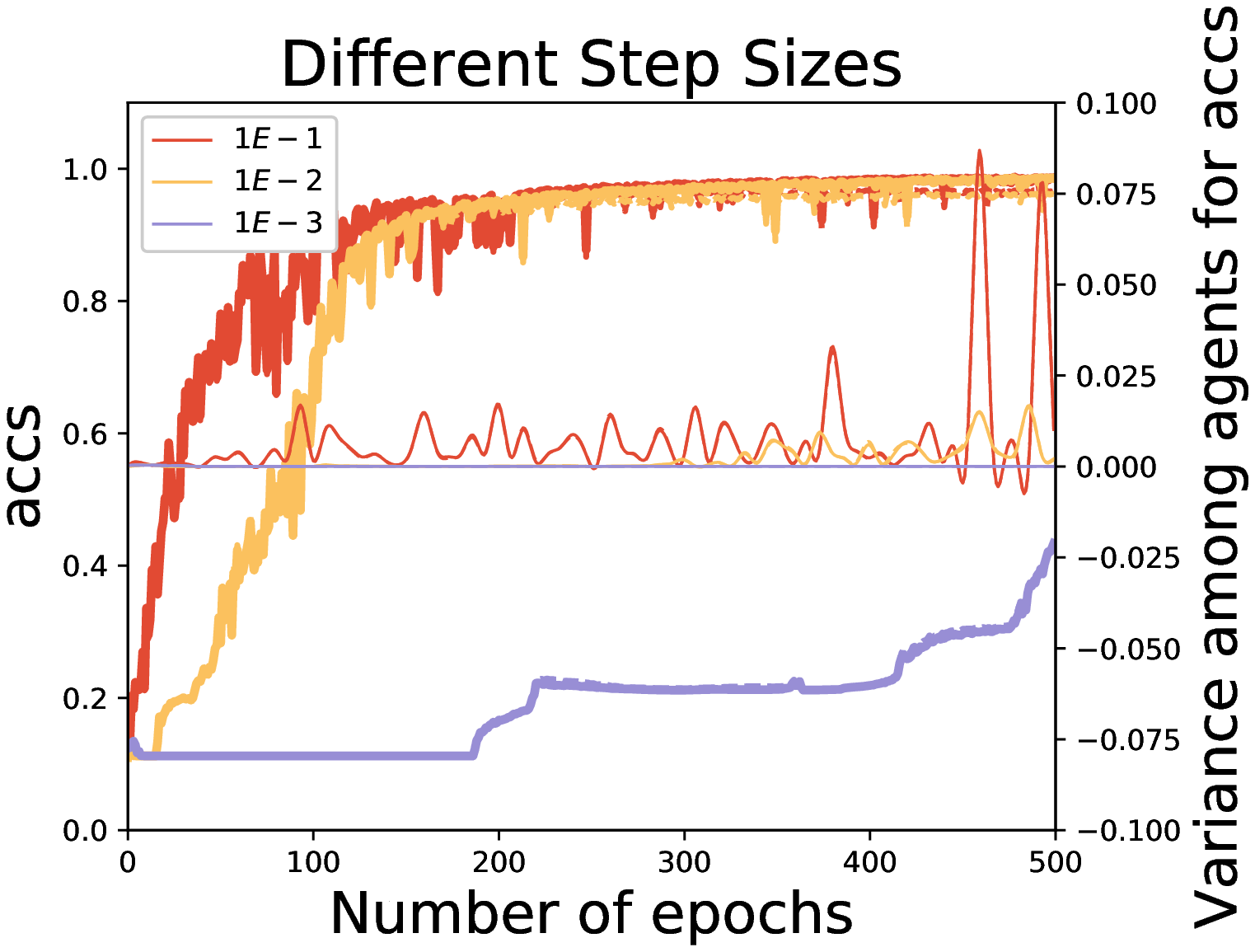}}

\caption{\textit{Average training (solid lines) and validation (dash lines) (a) loss and (b) accuracy for SGD, MSGD and CDMSGD method for the MNIST dataset for decaying step size.  (c) loss and (d) accuracy for CDMSGD for the MNIST data with different learning rates }}\label{Figure5}
\end{figure}

\subsubsection{Effect of step size}
The analysis presented in this paper shows that choice of step size is critical in terms of convergence as well as accuracy. To explore this aspect experimentally, we compare the performance of CDMSGD for three different fixed step sizes using MNIST data. The results are presented in ~\ref{Figure5} (c) \& (d), where the (fixed) step size was varied from $0.1 (1E-1)$ to $0.01 (1E-2)$ and then to $0.001 (1E-3)$. While the fastest convergence of the algorithm is observed with step size 0.1, the level of consensus (indicated by the variance among the agents) is quite unstable. On the other hand, with very low step size $0.001$, the level of consensus is quite stable (moving average of variance remains $0$). However, the convergence is extremely slow. This observation conforms to the theoretical analysis described in the paper as well as justifies the choice of step size $0.01$ in the experiments presented above. %This observation is trivial and follows the theoretical results presented above.

\end{document}